\documentclass[lettersize,journal]{IEEEtran}
\usepackage{color}
\usepackage{newtxtext,newtxmath}
\usepackage{amsmath}
\usepackage{graphicx}
\graphicspath{{figures/}}
\usepackage[caption=false]{subfig}
\usepackage{array}
\usepackage{booktabs}
\usepackage{multirow}
\usepackage{cite}
\usepackage{comment}
\usepackage[colorlinks,allcolors=blue]{hyperref}
\newcommand{\num}[1]{n_{#1}}
\newcommand{\pop}{\boldsymbol{P}}
\newcommand{\ind}{\boldsymbol{p}}
\newcommand{\offs}{\boldsymbol{Q}}
\newcommand{\combinedPop}{\boldsymbol{R}}
\newcommand{\front}{\boldsymbol{F}}
\newcommand{\subPop}{\boldsymbol{S}}
\newcommand{\objFunc}{f}
\newcommand{\imgW}{W}
\newcommand{\imgH}{H}
\newcommand{\tplImg}{I}
\newcommand{\inpImg}{I^\prime}
\newcommand{\tplCoord}{\boldsymbol{x}}
\newcommand{\inpCoord}{\boldsymbol{x}^\prime}
\newcommand{\spacing}{s}
\newcommand{\dispVec}{\boldsymbol{d}}
\newcommand{\dispElem}{d}
\newcommand{\deformation}{D}
\newcommand{\bspline}{B}
\newcommand{\xform}{T}

\begin{document}

\title{Image Deformation Estimation via Multiobjective Optimization}

\author{%
Takumi~Nakane,
Haoran~Xie,
and~Chao~Zhang
\thanks{This work was supported in part by the Japan Society for the Promotion of Science (JSPS) KAKENHI under Grant JP20K19568, and in part by the Support Center for Advanced Telecommunications Technology Research (SCAT) Grant.}
\thanks{Takumi Nakane is with the Department of Engineering, the University of Fukui, Fukui 910-8507, Japan.}
\thanks{Haoran Xie is with the Japan Advanced Institute of Science and Technology, Nomi 923-1211, Japan.}
\thanks{Chao Zhang is with the Department of Engineering, the University of Fukui, Fukui 910-8507, Japan (e-mail: zhang@u-fukui.ac.jp).}
}

\maketitle

\begin{abstract}
The free-form deformation model can represent a wide range of non-rigid deformations by manipulating a control point lattice over the image. However, due to a large number of parameters, it is challenging to fit the free-form deformation model directly to the deformed image for deformation estimation because of the complexity of the fitness landscape. In this paper, we cast the registration task as a multi-objective optimization problem (MOP) according to the fact that regions affected by each control point overlap with each other. Specifically, by partitioning the template image into several regions and measuring the similarity of each region independently, multiple objectives are built and deformation estimation can thus be realized by solving the MOP with off-the-shelf multi-objective evolutionary algorithms (MOEAs). In addition, a coarse-to-fine strategy is realized by image pyramid combined with control point mesh subdivision. Specifically, the optimized candidate solutions of the current image level are inherited by the next level, which increases the ability to deal with large deformation. Also, a post-processing procedure is proposed to generate a single output utilizing the Pareto optimal solutions. Comparative experiments on both synthetic and real-world images show the effectiveness and usefulness of our deformation estimation method.
\end{abstract}


\begin{figure*}
\centering
\includegraphics[width=\linewidth]{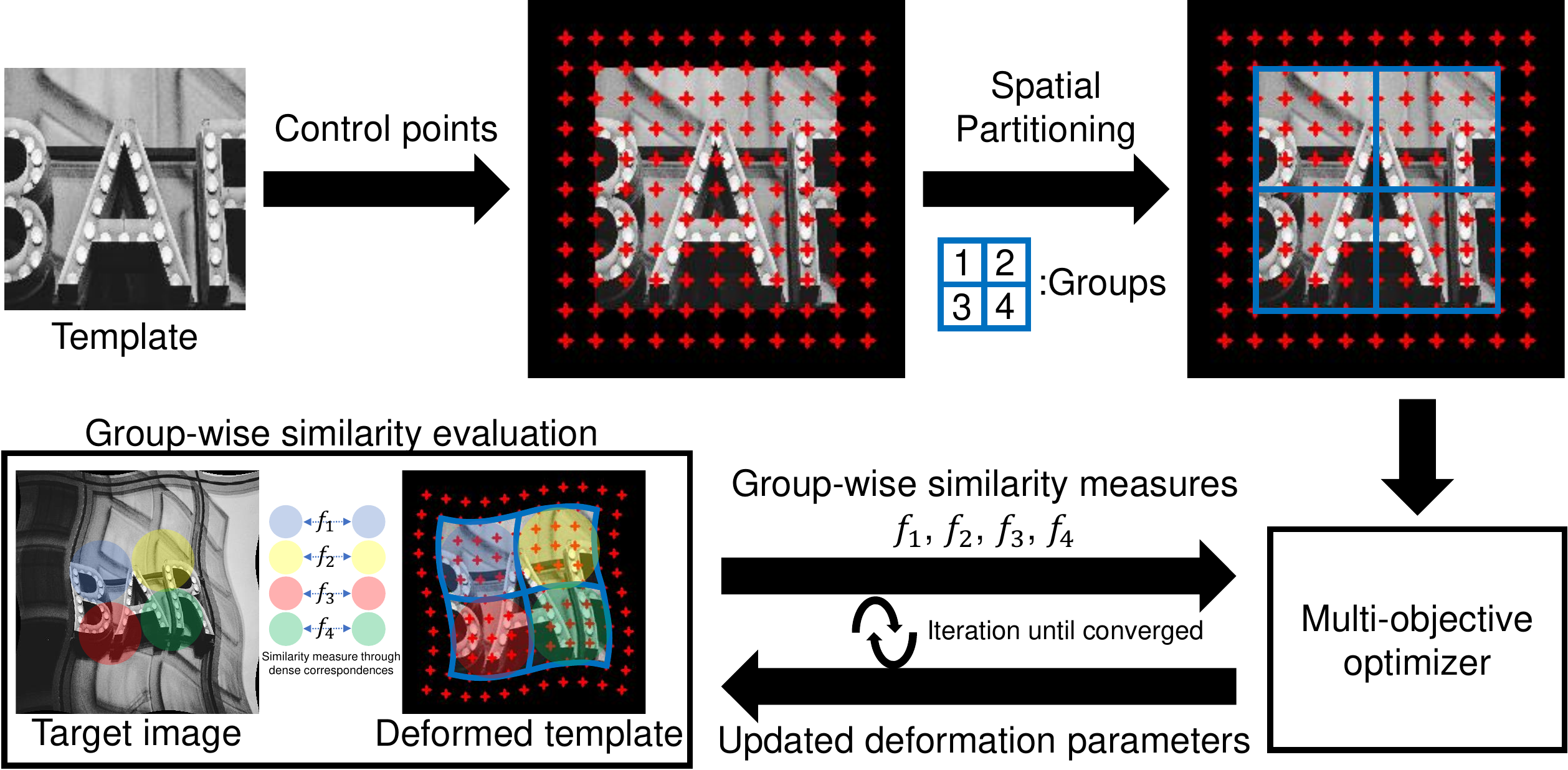}
\caption{Problem setting and overview (group number is set to four here). Image deformation estimation can be regarded as an optimization problem where the goal is to search the deformation parameters which maximize the similarity between the deformed template image and the target image.}
\label{fig:introduction_optimization}
\end{figure*}

\section{Introduction}
\label{sec:introduction}
Estimation of the deformation parameters of a target (either objects or texture) is a fundamental technique mainly for computer vision applications such as registration \cite{gay2010direct,cao2018deformable,balakrishnan2019voxelmorph} and tracking \cite{tan2014deformable,wang2019deformable,wang2020tracking}. If the geometric deformation model is constrained to only rotation and translation then the deformation is rigid. Affine or projective transformations can express more complex deformation, while practical applications, such as medical image analysis \cite{rueckert1999nonrigid,oliveira2014medical} and morphing \cite{alexa2002recent,scherhag2019face}, often involve non-rigid deformation with more degrees of freedom. Furthermore, the deformation between two images can be global, local, and even space-invariant, which makes the problem more challenging because the movement vector of each pixel is required to be estimated independently while preserving the smoothness. This technique can estimate the deformation parameters by deforming a template image such that the similarity between the deformed template and a target image is maximized. The procedure of similarity maximization can be cast as an optimization problem by treating deformation parameters as decision variables and the similarity as the objective function.

As to the geometric deformation model, free-form deformation (FFD) \cite{sederberg1986free} models the deformation by manipulating control points arranged in a regular lattice over the target. Each pixel moves based on weights by basis functions and displacements of surrounding control points. B-spline basis functions are generally used to weigh displacements \cite{hsu1992direct}. The larger the number of control points is, the more finely the deformation can be modeled. In FFD, the influence of a control point is limited to neighbor pixels, which brings benefits with respect to ability in modeling and computational cost \cite{crum2004non}. Due to these characteristics, FFD is allowed to model highly free and subtle deformation. However, optimization with FFD's parameters is challenging since an optimizer needs to treat the displacements of all control points as decision variables, and it is obvious that the expressive power of the deformation model is proportional to the number of parameters. Moreover, because each control point can affect multiple regions, an improvement of similarity in one region may negatively make similarity in other regions worse, i.e., there exist conflicts between regions.

To alleviate the above issues, we introduce a new idea to estimate the FFD parameters by casting the deformation estimation problem as a multi-objective optimization problem (MOP), which can be effectively solved by multi-objective evolutionary algorithms (MOEAs). The overview of our algorithm is shown in Fig. \ref{fig:introduction_optimization}. A template is spatially divided into several groups and the similarity measure over each group is treated as a single-objective function and independently computed. Each group consists of patches, and the pixels in each patch are affected by the same control points. A MOP requires simultaneous optimization of two or more objectives that conflict with each other. In our problem setting, we aim to find Pareto optimal solutions given none of the groups can be improved without degrading some of the other groups, which can be solved by certain off-the-shelf MOEAs. In addition, we adopt a coarse-to-fine strategy using image pyramids to improve the estimation capability, especially for large deformations. Specifically, the optimization starts at the top of the pyramid (i.e., the lowest resolution image) and is executed at each level of the pyramid. The number of control points is gradually increased as the resolution increases, and the interpolation of control points is realized by mesh subdivision to allow fine-grained deformation. Also, a post-processing method is proposed to integrate Pareto optimal solutions into a single output as the decision-making procedure. For each group in Fig. \ref{fig:introduction_optimization}, the group-wise deformation parameters with the highest group-wise similarity are adopted. These group-wise deformation parameters are aggregated into a final solution. We perform comparative experiments using both synthetic and real-world data to show the effectiveness and usefulness of our method. In conclusion, our contributions are threefold. 
\begin{itemize}
\item The deformation estimation problem is cast as a MOP by spatially dividing an image into multiple groups accompanied by independent similarity measures.
\item The estimation capability is improved by a coarse-to-fine strategy, which is realized by building image pyramids and conducting mesh subdivisions at each level.
\item A post-processing method is proposed to integrate Pareto optimal solutions into a final output.
\end{itemize}

The rest of this paper is organized as follows. We present related works of deformation estimation and MOEAs in Sec. \ref{sec:related} followed by a brief review of FFD model in Sec. \ref{sec:ffd}. The overview of three off-the-shelf genetic algorithms (GAs) are given in Sec. \ref{sec:ga}. In Sec. \ref{sec:proposed}, we describe the details of the spatial multi-objective problem and the coarse-to-fine optimization strategy. The experimental results are shown in Sec. \ref{sec:experiment}. Conclusion is given in Sec. \ref{sec:conclusion}.

\section{Related Work}
\label{sec:related}

\subsection{Deformation Estimation Between Two Images}
\label{sec:related_deformation}
Many methods have been proposed to deal with deformable surfaces, which can be roughly categorized as feature-based methods and pixel-based methods. The former category estimates deformation parameters by using feature correspondences commonly extracted from two images. The latter category maximizes the similarity calculated using dense pixels directly. Also, hybrid methods have been studied to incorporate the advantages of both approaches \cite{zhu2009fast,pizarro2012feature,wu2013multiple}.

Feature-based methods \cite{pilet2008fast,salzmann2011linear,tran2012defence,ngo2016template,wang2019deformable} estimated deformation parameters based on the correspondences between feature points between the template image and the target image. The accuracy largely depends on the quality of the correspondences. Therefore, the elimination of outliers from the extracted feature set is an essential process. However, the large number of parameters in free-form deformation makes it difficult to apply standard methods such as RANSAC \cite{tran2012defence}. Other limitations are: 1) In the case of feature-less images, feature points are hard to be detected. Without inlier correspondences, the parameters cannot be appropriately estimated. Especially in the case of non-rigid transformations, more corresponding points are required \cite{bartoli2004direct}. 2) Local features such as SIFT \cite{lowe2004distinctive} and ASIFT \cite{morel2009asift} are susceptible to complex transformation, which may largely degrade the confidence of correspondences when complex transformation occurs.

The purpose of pixel-based methods \cite{bartoli2004direct,gay2006image,malis2007efficient,hilsmann2010realistic,gay2010direct,tan2014deformable,zhang2015fast,zhang2016robust} is to solve the minimization problem of the cost function consisting of a data term and some restrictions such as smoothness term calculated from pixel intensities. The data term is usually defined as the sum of the intensity differences between the pixels of the template image and the corresponding pixels in the deformed target image. Such methods are less dependent on image features compared to feature-based methods. In addition, the capability of dealing with self-occlusion is a notable point. Since only a few features typically exist near the self-occlusion boundary, the pixel-based methods are more reasonable in such cases \cite{gay2010direct,pizarro2012feature}. In \cite{gay2010direct}, a penalty term called shrinker is incorporated into the cost function. The shrinker term acts to shrink the displacement in order to make self-occluded areas disappear. \cite{pizarro2012feature} employed a pixel-based approach to refine the deformation parameters given by a proposed feature-based method. When self-occlusion or strong deformations are involved, the hybrid method shows better results than only using the feature-based method. There also exist researches to maximize the similarity under the framework of evolutionary computation with a single-objective \cite{zhang2015fast,zhang2016robust}. There also exist methods achieving the minimization of the cost function by employing non-linear least squares solvers, such as the Gauss-Newton algorithm \cite{bartoli2004direct,gay2010direct,pizarro2012feature}, the Levenberg-Marquardt algorithm \cite{gay2006image,hilsmann2010realistic}, and the learning-based methods \cite{tan2014deformable}. To the best of our knowledge, exploiting evolutionary algorithms \cite{klein2007evaluation} or multi-objective optimization approaches \cite{alderliesten2012multi,pirpinia2019evolutionary} to deal with deformable surfaces have been sparsely treated so far. Our previous work appearing in GECCO2019 addressed this problem by using a modified single-objective GA \cite{nakane2019probabilistic}. Different from the previous work, in this paper, we attempt to adopt evolutionary algorithms for solving this problem by casting it as a multi-objective optimization problem.

\subsection{Multiobjective Evolutionary Algorithms (MOEAs)}
\label{sec:related_moea}
EAs are optimization algorithms inspired by Darwin's evolutionary theory, such as the GA, evolutionary strategy, and evolutionary programming. These algorithms share a common framework in which many candidate solutions are simultaneously dealt with and stochastic operations are iteratively applied. Because of the powerful exploration capability, EAs have been applied in a variety of computer vision tasks. Interested readers can also refer to the survey \cite{nakane2020application}. EAs are also effective tools for solving MOPs. The population-based search procedure provides the advantage of finding the Pareto optimal solutions in a single run. MOEAs use dominance relation to rank solutions in an objective space consisting of conflicting objectives. In particular, representative MOEAs, such as non-dominated sorting GA-II (NSGA-II) \cite{deb2002fast}, strength Pareto EA2 (SPEA2) \cite{zitzler2002spea}, and Pareto enveloped based selection algorithm-II (PEAS-II) \cite{corne2001pesa}, include a mechanism that preserves non-dominated solutions in every generation, called elitism, and hence these algorithms can outperform non-elitist MOEAs by preventing the loss of good solutions \cite{vachhani2015survey}. Since the goal of MOEAs is to provide solutions that are widely distributed on the Pareto front, MOEAs are also required to maintain the diversity of solutions. In NSGA-II, a crowding distance was proposed which is the sum of the distances between the two nearest solutions for each objective. SPEA2 used the inverse of the distance to the k-th nearest solution as the density. PEAS-II divided the objective space into several hyperboxes and counted the number of solutions within them. The density was assigned to each hyperbox as the number of solutions contained. On the other hand, MOEAs are less effective for problems with four or more objectives, i.e., many-objective optimization problems (MaOPs). The main reason is that as the number of objectives increases, the condition of dominance becomes more complex. More objectives lead to a greater proportion of non-dominated solutions, and hence the ability of convergence toward the Pareto front decreases \cite{garza2009ranking}. There are several strategies to adapt MOEAs to MaOPs \cite{li2015many}, such as dimensionality reduction \cite{saxena2013objective,wang2016objective} and use of indicators \cite{bader2011hype,li2016stochastic}. Among them, one representative strategy is via decomposition, e.g., MOEA based on decomposition (MOEA/D) \cite{zhang2007moea}, where a MaOP was decomposed into single-objective sub-problems using weighting vectors, and reference-point based many-objective NSGA-II (NSGA-III) \cite{deb2014evolutionary}, where an objective space was divided by reference vectors. There also exist various improved versions of MOEA/D \cite{xu2019moea,zhang2020enhancing} and NSGA-III \cite{yuan2016new,cui2019improved}. Combination of both merits is also shown in \cite{li2015evolutionary}.

\section{Deformation Model}
\label{sec:ffd}
\begin{figure}
\centering
\includegraphics[width=0.8\linewidth]{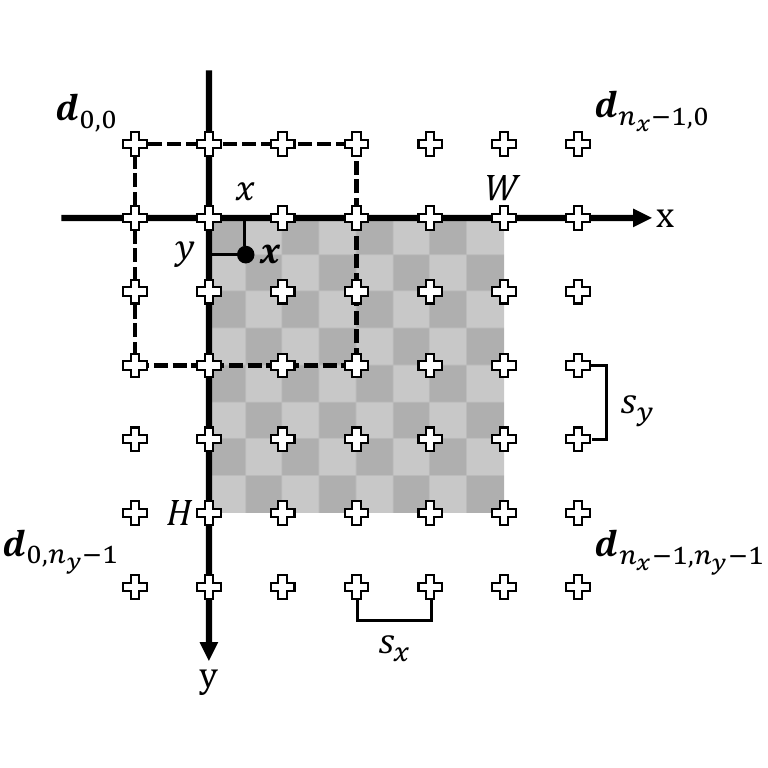}
\caption{The control point lattice configuration of FFD. Deformation of the pixel coordinate $\tplCoord$ is determined by the 16 control points within/on the dashed rectangle.}
\label{fig:ffd_configuration}
\end{figure}

Deformation estimation is achieved by registering the template image $\tplImg$ to the target image $\inpImg$. In order to deform $\tplImg$, we employ FFD combined with cubic B-splines using control point meshes. For $\tplImg$ of $\imgW \times \imgH$ pixels, control points are arranged on the $\num{x} \times \num{y}$ lattice with horizontal spacing $\spacing_x = \lceil \imgW / (\num{x} - 3) \rceil$ and vertical spacing $\spacing_y = \lceil \imgH / (\num{y} - 3) \rceil$ (i.e., the outmost control points are outside the region of $\tplImg$), as illustrated in Fig. \ref{fig:ffd_configuration}. Each control point is assigned a displacement vector $\dispVec$ representing the distance and direction from the initial position, and the movement of a certain coordinate $\tplCoord = (x, y)^\top$ on $\tplImg$ is determined by surrounding control points, which is defined as:
\begin{equation}
\deformation(\tplCoord) = \sum_{m=0}^3 \sum_{n=0}^3 \bspline_m(u) \bspline_n(v) \dispVec_{i + m, j + n},
\label{equ:ffd}
\end{equation}
where $i = \lfloor x / \spacing_x \rfloor, j = \lfloor y / \spacing_y \rfloor, u = x / \spacing_x - i, v = y / \spacing_y - j$, and $\bspline_m, \bspline_n$ are the cubic B-spline basis functions \cite{lee1997scattered}. Then, the pixel coordinate $\inpCoord$ on $\inpImg$ corresponding to $\tplCoord$ is given by the transformation function $\xform$:
\begin{equation}
\inpCoord = \xform(\tplCoord) = \tplCoord + \deformation(\tplCoord).
\label{equ:xform}
\end{equation}

\begin{figure}
\centering
\subfloat[]{
    \includegraphics[width=0.45\linewidth]{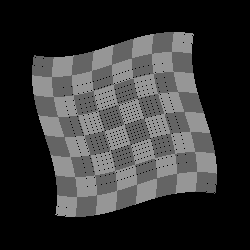}
    \label{fig:ffd_example_forward}
}\quad
\subfloat[]{
    \includegraphics[width=0.45\linewidth]{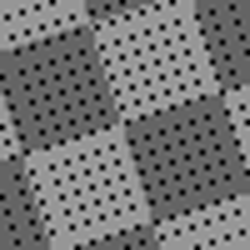}
    \label{fig:ffd_example_extended_forward}
}\\
\subfloat[]{
    \includegraphics[width=0.45\linewidth]{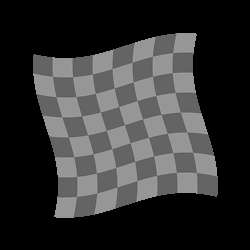}
    \label{fig:ffd_example_backward}
}\quad
\subfloat[]{
    \includegraphics[width=0.45\linewidth]{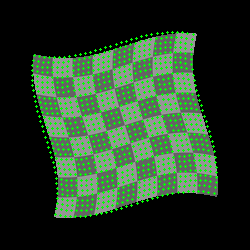}
    \label{fig:ffd_example_difference}
}
\caption{(a) Deformed Fig. \ref{fig:ffd_configuration} by forward warping. (b) A closeup view around the center of (a). (c) Backward warping. (d) Shape comparison between forward warping (green dots are sampled points) and backward warping.}
\label{fig:ffd_example}
\end{figure}

Since the transformation using Eq. \ref{equ:xform} is a forward warping procedure and $\inpCoord$ consists of real numbers, there is a problem that rounding operation is necessary to obtain pixel intensities. Such a process results in a large number of ``holes'' in the deformed template image, as illustrated in Fig. \ref{fig:ffd_example_forward} and Fig. \ref{fig:ffd_example_extended_forward}. An alternative is to employ backward warping that $\tplCoord$ corresponding to $\inpCoord$ is computed using the inverse transformation $\xform^{-1}$ and interpolation scheme can be used to obtain the pixel intensity. According to \cite{schwarz2007non}, $\xform^{-1}$ can be defined using an approximation:
\begin{equation}
\tplCoord = \xform^{-1}(\inpCoord) \approx \inpCoord - \deformation(\inpCoord).
\label{equ:inv_xform}
\end{equation}

The deformation obtained in Eq. \ref{equ:inv_xform} and its difference from Eq. \ref{equ:xform} are illustrated in Fig. \ref{fig:ffd_example_backward} and Fig. \ref{fig:ffd_example_difference}, respectively. In conclusion, Eq. \ref{equ:ffd} is employed for modeling geometric deformation and Eq. \ref{equ:inv_xform} is employed as the image transformation for registration in this paper.

\section{Review of Genetic Algorithm (GA)}
\label{sec:ga}
To further explain how we use the multi-objective optimizer in Fig. \ref{fig:introduction_optimization} to solve the optimization problem, we provide a short review on GAs in this section. GA is one of the leading algorithms in nature-inspired optimization methods. For a population $\pop$ consisting of a number of candidate solutions, the GA gradually optimizes the $\pop$ by iteratively applying genetic operators. One of the unique features of the GA is genotype representations for candidate solutions. Each candidate solution, called individual $\ind$, is encoded into an internal representation, such as a bit string or a real-valued vector, in order to apply genetic operators. The genotype representations can allow genetic operators to adapt to different problems flexibly.

The genetic operators mainly consist of parents selection, crossover and mutation. The procedure of the simple GA \cite{holland1975adaptation} is briefly listed as follows:
\begin{description}
\item[Step 1:] Set $t=0$ and generate $\num{\pop}$ individuals in $\pop_t$ randomly.
\item[Step 2:] Select individuals as parents from $\pop_t$ with parents selection operator.
\item[Step 3:] Generate an offspring population $\offs_t$ from parents by crossover and mutation operators.
\item[Step 4:] Evaluate all individuals in $\offs_t$.
\item[Step 5:] Select $\num{\pop}$ individuals from $\offs_t$ as $\pop_{t+1}$.
\item[Step 6:] Set $t=t+1$ and return to Step 2 until the termination criterion is satisfied.
\end{description}
Parent selection is typically implemented as probabilistic selection biased by evaluation values (i.e., individuals with better evaluation values are assigned higher probabilities.) The crossover operation generates offspring through the genetic recombination of multiple parents. The mutation operation randomly changes genes of offspring with low probability. Step 5 is also known as survivors selection, that is, all individuals in $\pop_t$ and $\offs_t$ compete in order to become members of $\pop_{t+1}$ according to their evaluation values. The simple GA directly adopts $\offs_t$ as $\pop_{t+1}$. Besides, the elitism strategy, which preserves the best individual in the pool $\combinedPop_t=\pop_t \cup \offs_t$ is often adopted.

NSGA-II \cite{deb2002fast} is a representative GA-based algorithm for solving MOPs. The key idea of the NSGA-II is to introduce the selection criterion using two sorting approaches. The first one, called fast non-dominated sorting, iteratively extracts a non-dominated set $\front$ from the population and assigns a rank to each $\front$ according to the order in which they are extracted. Another one, called crowding distance sorting, determines the priority in $\front$ by the crowding distance which represents the density of neighboring individuals in the solution space. At last, $\num{\pop}$ individuals are selected from $\subPop$ consisting of $\front_1$ to $\front_{l-1}$, where $\vert \subPop \vert \leq \num{\pop} < \vert \subPop \vert + \vert \front_l \vert$. Insufficient individuals are taken from $\front_l$ according to the crowding distance. Fast non-dominated sorting promotes convergence to the Pareto front, while crowding distance sorting maintains diversity on the Pareto front. Moreover, elitism is ensured by using $\combinedPop$ in survivors selection.

NSGA-III \cite{deb2014evolutionary} is a variant of NSGA-II which focuses on solving MaOPs. Instead of crowding distance sorting, reference lines connecting the origin with reference points evenly distributed on the evaluation value space are used to maintain the diversity of the population on the Pareto front. Each $\ind$ is associated with the closest reference line in the perpendicular distance. After $\subPop$ is determined by fast non-dominated sorting in survivors selection, the number of individuals in $\subPop$ associated with each reference line, called niche count, is logged. NSGA-III then iteratively selects the individual in $\front_l$ which is associated with the reference line of the lowest niche count. Reference points can relieve the algorithm in adaptively maintaining population diversity. In addition, users can obtain only a part of the Pareto front as required by manually distributing reference points.

\section{Spatial Multiobjective Optimization}
\label{sec:proposed}
\begin{figure}
\centering
\includegraphics[width=\linewidth]{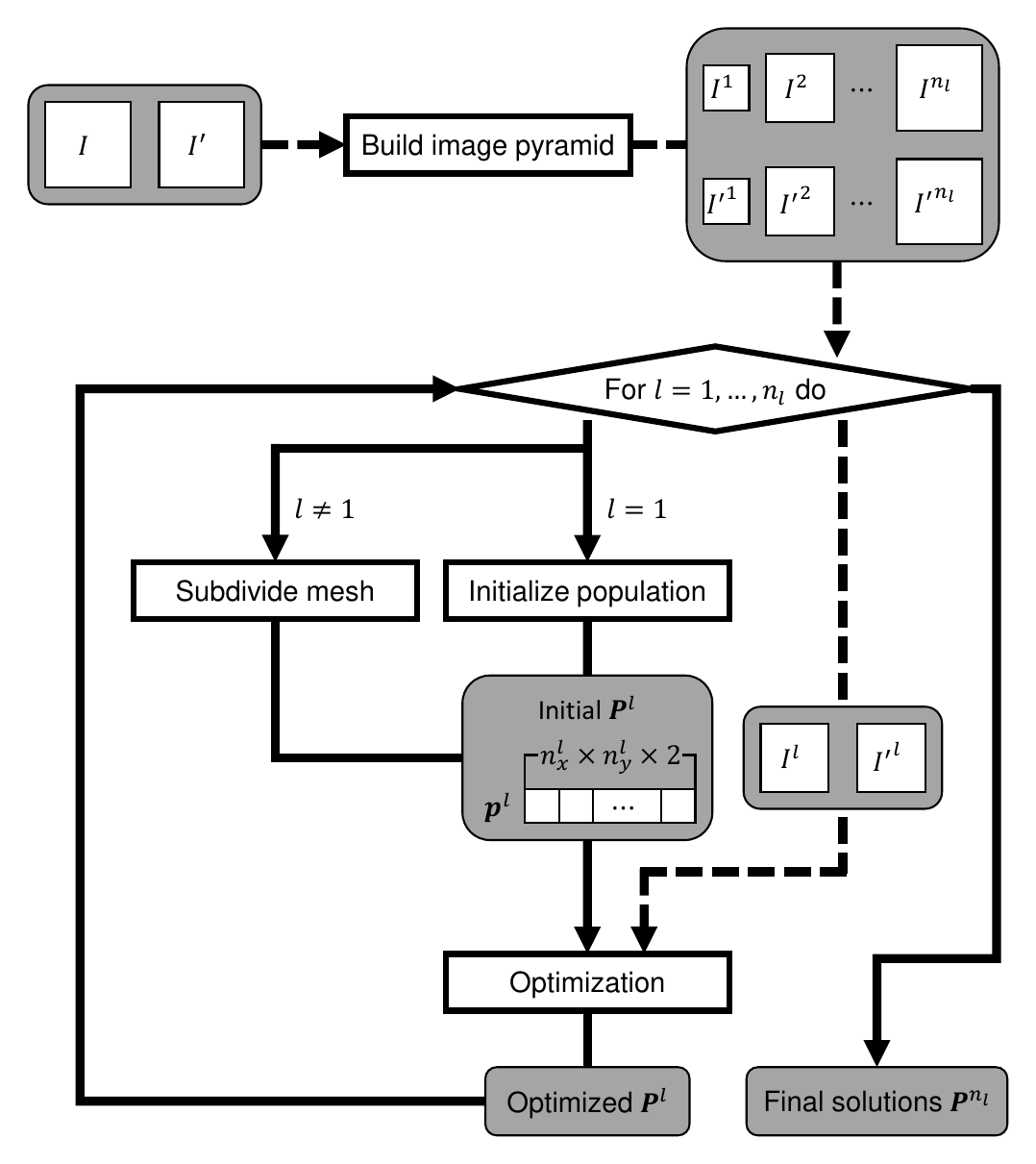}
\caption{Overview of the optimization procedure.}
\label{fig:overview}
\end{figure}

As described in Sec. \ref{sec:ffd}, the deformation of the template image is determined by the displacement of each control point. Therefore, the purpose of the optimization procedure is to calculate the displacements where the deformed template image matches the target in the target image most. The principal contribution of this work is to cast this task as a MOP by spatially partitioning the template into groups. Each group is assigned a single-objective function of the similarity measure. The overview of the optimization procedure is shown in Fig. \ref{fig:overview}. The procedure starts from building pyramids for both the template image and the target image, then the optimization is performed with the pyramids in a coarse-to-fine scheme. The key advantage of this framework is that the population optimized at each level can be inherited as the initial population of the next level. To ensure the consistency of parameter inheritance from low-resolution level to high-resolution level, a subdivision method considering control point mesh is employed to achieve a natural interpolation of additional displacements, which allows the optimized population to be directly inherited.

Final candidate solutions are obtained by optimizing the population at the bottom of the pyramid (i.e., the image in the original resolution). The subsequent effort lies in how to determine a final solution as output from multiple solutions on the Pareto front. In addition to a direct selection approach, a post-processing approach using multiple candidate solutions is also proposed.

\subsection{Optimizing Deformation Parameters via MOEAs}
\label{sec:proposed_optimization}
Three GAs described in Sec. \ref{sec:ga} are employed and compared to optimize the displacements in the experiment. Since each control point can move within the plane, the total number of decision variables is $2\num{x}\num{y}$. To represent these variables as genes, we use real-valued coding because each displacement is a 2D vector of real values. For clarity, we denote the displacement of a single point by $\dispVec_{i, j} = (\dispElem_{i, j, x}, \dispElem_{i, j, y})^\top$, and an individual $\ind$ is represented by the vector concatenating the displacements of all the points as follows,
\begin{equation}
\ind = (\dispElem_{0, 0, y}, \dispElem_{0, 0, x}, \dispElem_{1, 0, y}, \ldots, \dispElem_{\num{x} - 1, \num{y} - 1, x})^\top.
\label{equ:individual}
\end{equation}
$\ind$ can directly represent a candidate control point mesh. The initial population is iteratively optimized by genetic operators. As to the evaluation of each $\ind$, for simplicity, a single-objective function is firstly introduced, which is combined with the simple GA and compared to multi-objective GAs in the experiment. In the objective function, mean absolute difference with respect to intensities is used as the similarity measure. As introduced in Sec. \ref{sec:ffd}, we use backward warping to find the correspondences for the calculation of objective function, hence sampling is performed on the target image. Let $\Omega$ and $\Omega^\prime$ denote the entire region of the template image $\tplImg$ and the target image $\inpImg$, respectively, and we can further define the sampling region $\omega^\prime \subseteq \Omega^\prime$ as:
\begin{equation}
\omega^\prime = \{\inpCoord \mid \xform^{-1}(\inpCoord) \in \Omega \}.
\label{equ:sample_region}
\end{equation}
The single-objective function for the simple GA is given by:
\begin{equation}
\objFunc(\ind) = \frac{1}{\vert \omega^\prime \vert} \sum_{\inpCoord \in \omega^\prime} \vert \inpImg(\inpCoord) - \tplImg(\xform^{-1}(\inpCoord)) \vert.
\label{equ:sobj}
\end{equation}
Based on Eq. \ref{equ:sobj}, spatial multi-objective functions can be easily defined. We first define \textit{patch} as a region consisting of pixels affected by the shared $4\times4$ control points, i.e., $\Omega$ consists of $(\num{x} - 3) \times (\num{y} - 3)$ patches. The group partitioning is achieved by dividing all the patches into $\num{\objFunc}$ groups, where $\num{\objFunc}$ is the number of objective functions. We denote the region of $i$-th group as $\omega_i \subseteq \Omega$. One objective function is assigned over each group, which can be written by modifying Eq. \ref{equ:sample_region} and Eq. \ref{equ:sobj}:
\begin{equation}
\omega_i^\prime = \{\inpCoord \mid \xform^{-1}(\inpCoord) \in \omega_i \},
\label{equ:sample_region_i}
\end{equation}
\begin{equation}
\objFunc_i(\ind ) = \frac{1}{\vert \omega_i^\prime \vert} \sum_{\inpCoord \in \omega_i^\prime} \vert \inpImg(\inpCoord) - \tplImg(\xform^{-1}(\inpCoord)) \vert,
\label{equ:mobj_i}
\end{equation}
where $i = 1, 2, \ldots, \num{\objFunc}$. Therefore, multi-objective GAs evaluate individuals based on the following vector function:
\begin{equation}
\boldsymbol{\objFunc}(\ind) = (\objFunc_1(\ind), \objFunc_2(\ind), \ldots, \objFunc_{\num{\objFunc}}(\ind))^\top.
\label{equ:mobj}
\end{equation}
Because each control point affects multiple patches, their similarity functions can ``conflict'' with each other, which is also considered in the case of groups consisting of multiple patches. This is an important motivation for adopting MOEAs because a single-objective optimizer can hardly solve such a conflicting problem efficiently \cite{deb2002fast}. The number of regions is adjustable as a hyper-parameter (increasing $\num{\objFunc}$ makes optimization significantly more difficult). Pareto optimal solutions are supposed to obtain more appropriate solutions than optimizing a single-objective.

\subsection{Coarse-to-Fine Strategy}
\label{sec:proposed_ctof}
An iterative framework using image pyramids can be employed to alleviate the difficulty of deformation estimation with large displacements \cite{hilsmann2010realistic,gay2010direct}. Estimation starts from the lowest resolution image for a rough estimation, and more accurate estimations are achieved as the image resolution increases. Specifically, we adopt Gaussian pyramid which iteratively generates low resolution images through Gaussian smoothing. The assignment of pyramid level indices follows the order in which estimation is performed (i.e., the first and the $\num{l}$-th level images are with the lowest and the highest resolution, respectively). Both the image width and height are halved and hence the $l$-th level image has $1/4$ resolution of the $(l+1)$-th level image.

\begin{figure}
\centering
\includegraphics[width=0.8\linewidth]{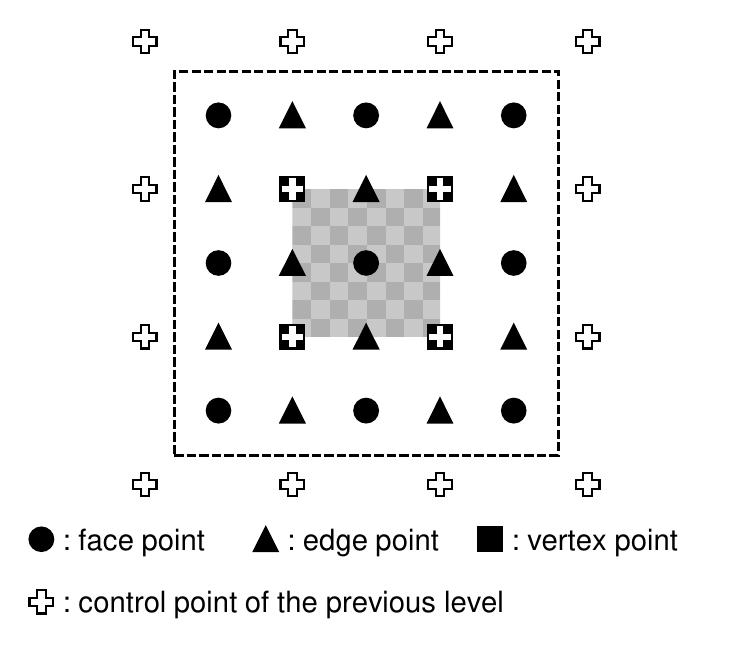}
\caption{Interpolation of control points by the subdivision algorithm for the current level. The middle image is the previous level and the control points of the current level (black) are computed based on the previous level's ones (white). The outermost control points (outside the dashed rectangle) are discarded and will not be involved in deformation. Black points will be further displaced to reflect the image deformation.}
\label{fig:subdivision}
\end{figure}

\begin{figure}
\centering
\subfloat[]{
    \includegraphics[width=0.3\linewidth]{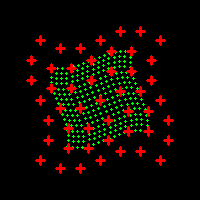}
    \label{fig:subdivision_example_previous}
}\quad
\subfloat[]{
    \includegraphics[width=0.6\linewidth]{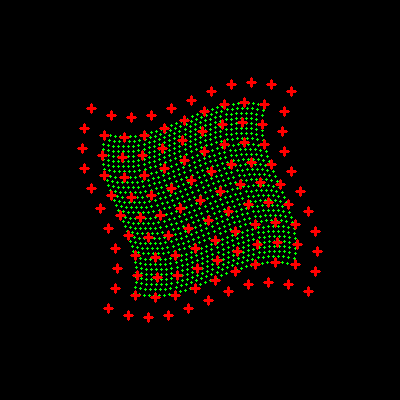}
    \label{fig:subdivision_example_current}
}
\caption{An example of a control point subdivision. Red markers represent the control points, and green dots represent the deformed surface with sampled points. (a) An example mesh with $7 \times 7$ lattice. (b) The subdivided mesh and enlarged surface with $11 \times 11$ lattice.}
\label{fig:subdivision_example}
\end{figure}

With the increase of image resolution, it is also necessary to increase the resolution of the control point mesh to inherit deformation parameters between different levels. To this end, for a certain level, interpolating new control points without destroying the mesh configuration of the previous level is required. In this work, we adopt the Catmull-Clark subdivision \cite{catmull1978recursively} for the mesh subdivision. The purpose of subdivision is to update the $\num{x}^l \times \num{y}^l$ control point mesh with respect to each individual from the optimized $\pop^l$ to $\num{x}^{l+1} \times \num{y}^{l+1}$ mesh, where $\num{x}^{l+1} = 2\num{x}^l - 3$ and $\num{y}^{l+1} = 2\num{y}^l - 3$ (i.e., $\spacing_x$ and $\spacing_y$ are fixed for all levels). The Catmull-Clark subdivision algorithm generates a subdivided mesh by inserting new control points and updating the existing control points. As illustrated in Fig. \ref{fig:subdivision}, the points on the subdivided mesh can be classified into three types: face points, edge points, and vertex points. A face point is inserted to a patch. Assuming that the vertices of a patch are $\dispVec_1^l, \dispVec_2^l, \dispVec_3^l,$ and $\dispVec_4^l$, the face point $\dispVec_F^{l+1}$ is computed as their centroid,
\begin{equation}
\dispVec_F^{l+1} = \frac{1}{4} \sum_{i=1}^{4} \dispVec_i^l.
\label{equ:face}
\end{equation}
An edge point is inserted to the edge shared by two patches. Assuming that two face points are $\dispVec_{F1}^{l+1}$ and $\dispVec_{F2}^{l+1}$, and two endpoints of the edge are $\dispVec_1^l$ and $\dispVec_2^l$, the edge point $\dispVec_{E}^{l+1}$ is computed as follow:
\begin{equation}
\dispVec_E^{l+1} = \frac{\dispVec_{F1}^{l+1} + \dispVec_{F2}^{l+1} + \dispVec_1^l + \dispVec_2^l}{4}.
\label{equ:edge}
\end{equation}
A vertex point is the updated point of a vertex $\dispVec^l$ shared by the four patches. Denoting that the average of the four face points is $\bar{\dispVec}_F$, and the average of the midpoints of the four edges which share $\dispVec^l$ as one of the endpoints is $\bar{\dispVec}_M$, the vertex point $\dispVec_{V}^{l+1}$ is computed as follows:
\begin{equation}
\dispVec_{V}^{l+1} = \frac{1}{4} \bar{\dispVec}_F + \frac{1}{2} \bar{\dispVec}_M + \frac{1}{4} \dispVec^l.
\label{equ:vertex}
\end{equation}
Note that the edge points and vertex points outside the dashed rectangle are not calculated, as surrounding control points are needed for calculation. After interpolation of control points, all the displacements are doubled to fit the increase of resolution in the image pyramid. An example of the subdivision process is shown in Fig. \ref{fig:subdivision_example}. It can be observed that the subdivided mesh in Fig. \ref{fig:subdivision_example_current} can maintain the shape of the previous mesh in Fig. \ref{fig:subdivision_example_previous} well. Therefore, this subdivision step is useful for providing a good initialization for optimizing the image of the next level in the pyramid.

\subsection{Decision of the Final Output}
\label{sec:proposed_decision}
\begin{figure}
\centering
\includegraphics[width=0.6\linewidth]{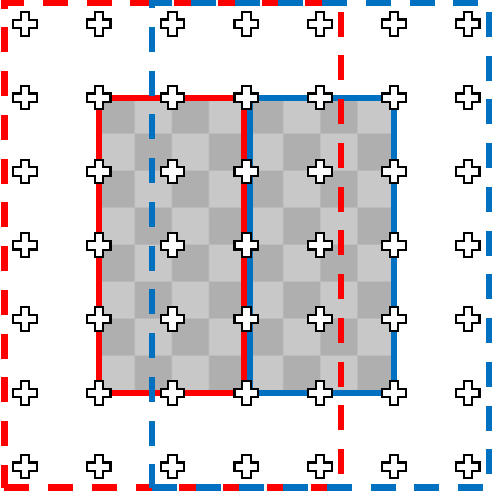}
\caption{An example of control point aggregation under 7$\times$7 lattice with two groups. The control points surrounded by one dashed rectangle are determined by the optimum solution which achieves the maximum similarity with respect to the group region surrounded by the solid rectangle of the same color. For shared control points (three columns in the center), the average of their displacements is computed.}
\label{fig:post_processing}
\end{figure}

We introduce a post-processing procedure to decide the final single solution as the output from the optimized population. A natural idea is to define the best solution as the individual with the smallest sum of the objective function values (i.e., maximum similarity). However, in MOEAs, such an approach can lose most of the valuable information of the Pareto optimal solutions. We propose a post-processed solution by exploiting these sub-optimal solutions, as illustrated in Fig. \ref{fig:post_processing}. For each group, the solution with the smallest value of the corresponding objective function is extracted, which provides the control points that only affect the corresponding group. The final output is created by aggregating control points provided from all the groups. For shared control points, the average of their displacements is computed.

\section{Experimental Results}
\label{sec:experiment}
The effectiveness and usefulness of solving the deformation estimation problem with the multi-objective scheme are verified using both synthetic data (Sec. \ref{sec:experiment_synthetic}) and real-world images (Sec. \ref{sec:experiment_real}). We compared the following four settings: the simple GA with a single-objective, NSGA-II and NSGA-III with two-objectives respectively, and NSGA-III with four-objectives. In the two-objective setting, patches are divided equally and vertically into two groups (e.g., for a $7 \times 7$ lattice including $4 \times 4$ patches, each group consists of $2 \times 4$ patches). Similarly, patches are divided vertically and horizontally into four groups in the four-objective setting (e.g., each group consists of $2 \times 2$ patches for a $7 \times 7$ lattice). The number of pyramid levels $\num{l}$ is fixed to three in all experiments. To reduce the computational cost, pixel sampling is performed for individual evaluation by scanning the target image at five pixel intervals. For the implementation of three GAs, the Platypus package\footnote{\url{https://github.com/Project-Platypus/Platypus}} is used, which is an evolutionary computation framework in Python which includes many MOEAs. To ensure the correctness and fairness of the experiment, we only manually set several essential parameters and fix other parameters following the default setting throughout the experiment. Specifically, the number of evaluations is set to 10000 and the number of reference points of NSGA-III is set to 100 for the two-objective setting and 120 for the four-objective setting. For fair comparisons, all experiments are executed five times with different random seeds considering probabilistic operations. The initial population $\pop^1_0$ for all settings is kept the same with respect to each random seed.

\begin{figure}
\centering
\subfloat[]{
    \includegraphics[width=0.45\linewidth]{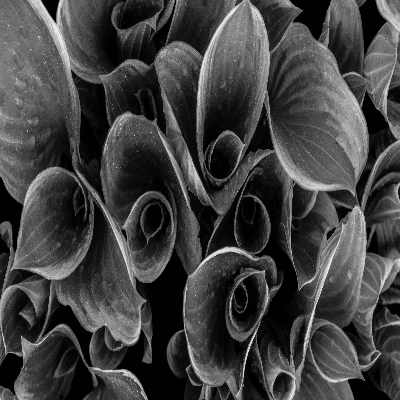}
    \label{fig:base_image_plant}
}\quad
\subfloat[]{
    \includegraphics[width=0.45\linewidth]{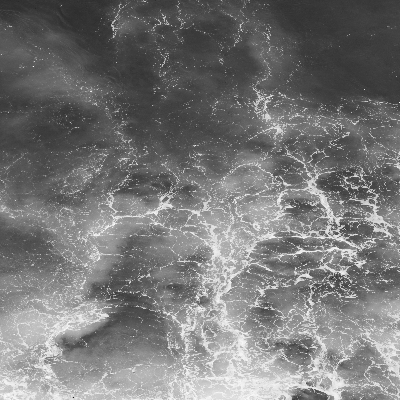}
    \label{fig:base_image_sea}
}\\
\subfloat[]{
    \includegraphics[width=0.45\linewidth]{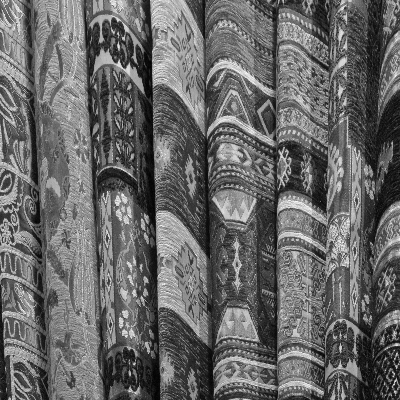}
    \label{fig:base_image_rag}
}\quad
\subfloat[]{
    \includegraphics[width=0.45\linewidth]{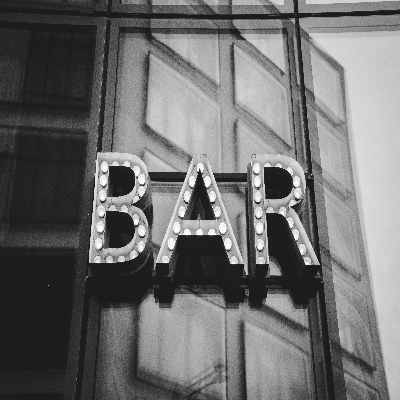}
    \label{fig:base_image_alphabet}
}\\
\subfloat[]{
    \includegraphics[width=0.45\linewidth]{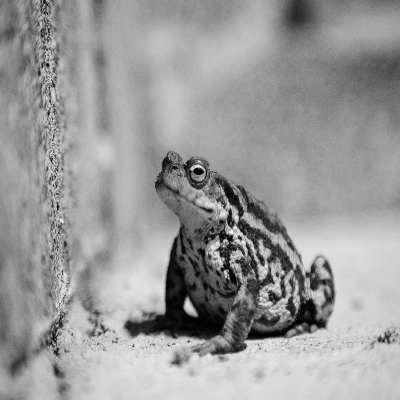}
    \label{fig:base_image_toad}
}
\caption{Images used for generating template and synthetically deformed target images: (a) plant, (b) sea, (c) rag, (d) alphabet, (e) toad.}
\label{fig:base_image}
\end{figure}

\begin{table*}[!t]
\centering
\caption{Comparative results of the best solutions with respect to the \textit{vertical wavy deformation}. Each cell consists of three values which represent the minimum, maximum, and average values based on five random trials from top to bottom. The best results in terms of RMSE and MEDE with respect to each row are emphasized in bold.}
\label{tab:vwave_best}
\scriptsize
\begin{tabular}{cccrrrrrrrr}
    \toprule
      &   &   & \multicolumn{4}{c}{RMSE} & \multicolumn{4}{c}{MEDE} \\
    \multirow{3}{*}{Image} & \multirow{3}{*}{\begin{tabular}{c} Lattice \\ size \end{tabular}} & Decision & \multicolumn{1}{c}{\multirow{3}{*}{GA}} & \multicolumn{1}{c}{Two} & \multicolumn{1}{c}{Two} & \multicolumn{1}{c}{Four} & \multicolumn{1}{c}{\multirow{3}{*}{GA}} & \multicolumn{1}{c}{Two} & \multicolumn{1}{c}{Two} & \multicolumn{1}{c}{Four} \\
      &   & variable &   & \multicolumn{1}{c}{objective} & \multicolumn{1}{c}{objective} & \multicolumn{1}{c}{objective} &   & \multicolumn{1}{c}{objective} & \multicolumn{1}{c}{objective} & \multicolumn{1}{c}{objective} \\
      &   & range &   & \multicolumn{1}{c}{NSGA-II} & \multicolumn{1}{c}{NSGA-III} & \multicolumn{1}{c}{NSGA-III} &   & \multicolumn{1}{c}{NSGA-II} & \multicolumn{1}{c}{NSGA-III} & \multicolumn{1}{c}{NSGA-III} \\
    \midrule
    \midrule
    \multirow{12}{*}{Plant} & \multirow{6}{*}{$7 \times 7$} & \multirow{3}{*}{$[-5.0, 5.0]$} & 2.28E+00 & \textbf{2.01E+00} & 2.16E+00 & 2.59E+00 & 1.22E-01 & \textbf{1.01E-01} & 1.02E-01 & 1.23E-01 \\
      &   &   & 3.00E+00 & 2.76E+00 & \textbf{2.53E+00} & 3.36E+00 & 1.51E-01 & \textbf{1.17E-01} & 1.24E-01 & 1.83E-01 \\
      &   &   & 2.64E+00 & \textbf{2.30E+00} & 2.34E+00 & 2.93E+00 & 1.35E-01 & \textbf{1.10E-01} & 1.13E-01 & 1.43E-01 \\ \cline{4-11}
      &   & \multirow{3}{*}{$[-10.0, 10.0]$} & 5.01E+00 & \textbf{4.30E+00} & 4.72E+00 & 5.05E+00 & 2.65E-01 & \textbf{2.13E-01} & 2.55E-01 & 2.76E-01 \\
      &   &   & 6.06E+00 & \textbf{5.55E+00} & 6.15E+00 & 6.90E+00 & 3.27E-01 & \textbf{3.02E-01} & 3.02E-01 & 3.52E-01 \\
      &   &   & 5.56E+00 & \textbf{4.92E+00} & 5.16E+00 & 5.93E+00 & 2.94E-01 & \textbf{2.62E-01} & 2.68E-01 & 3.11E-01 \\ \cline{4-11}
      & \multirow{6}{*}{$11 \times 11$} & \multirow{3}{*}{$[-5.0, 5.0]$} & 3.26E+00 & 2.73E+00 & \textbf{2.44E+00} & 3.26E+00 & 1.62E-01 & 1.48E-01 & \textbf{1.14E-01} & 1.66E-01 \\
      &   &   & 4.50E+00 & 4.23E+00 & \textbf{3.78E+00} & 4.16E+00 & 2.31E-01 & 2.09E-01 & \textbf{1.95E-01} & 2.14E-01 \\
      &   &   & 3.85E+00 & 3.34E+00 & \textbf{3.16E+00} & 3.65E+00 & 1.96E-01 & 1.79E-01 & \textbf{1.63E-01} & 1.92E-01 \\ \cline{4-11}
      &   & \multirow{3}{*}{$[-10.0, 10.0]$} & 6.12E+00 & \textbf{5.10E+00} & 5.16E+00 & 6.53E+00 & 3.35E-01 & \textbf{2.82E-01} & 3.42E-01 & 3.66E-01 \\
      &   &   & 9.34E+00 & 9.68E+00 & \textbf{7.65E+00} & 8.51E+00 & 6.91E-01 & 7.22E-01 & \textbf{5.21E-01} & 5.32E-01 \\
      &   &   & 7.37E+00 & 6.89E+00 & \textbf{6.84E+00} & 7.43E+00 & 4.67E-01 & 4.53E-01 & \textbf{4.41E-01} & 4.64E-01 \\
    \midrule
    \multirow{12}{*}{Sea} & \multirow{6}{*}{$7 \times 7$} & \multirow{3}{*}{$[-5.0, 5.0]$} & 3.79E+00 & 3.85E+00 & \textbf{3.49E+00} & 4.13E+00 & 1.03E-01 & 1.07E-01 & \textbf{9.17E-02} & 1.41E-01 \\
      &   &   & 4.19E+00 & \textbf{4.16E+00} & 4.36E+00 & 4.93E+00 & 1.28E-01 & \textbf{1.20E-01} & 1.24E-01 & 1.57E-01 \\
      &   &   & 3.97E+00 & 4.04E+00 & \textbf{3.94E+00} & 4.50E+00 & 1.18E-01 & 1.15E-01 & \textbf{1.13E-01} & 1.48E-01 \\ \cline{4-11}
      &   & \multirow{3}{*}{$[-10.0, 10.0]$} & 8.08E+00 & \textbf{7.68E+00} & 8.05E+00 & 9.10E+00 & \textbf{2.22E-01} & 2.26E-01 & 2.46E-01 & 2.80E-01 \\
      &   &   & 1.06E+01 & \textbf{9.11E+00} & 1.02E+01 & 1.04E+01 & 3.37E-01 & \textbf{2.58E-01} & 3.47E-01 & 3.73E-01 \\
      &   &   & 9.10E+00 & \textbf{8.46E+00} & 9.11E+00 & 9.57E+00 & 2.66E-01 & \textbf{2.41E-01} & 3.00E-01 & 3.16E-01 \\ \cline{4-11}
      & \multirow{6}{*}{$11 \times 11$} & \multirow{3}{*}{$[-5.0, 5.0]$} & 4.96E+00 & 4.19E+00 & \textbf{3.84E+00} & 4.71E+00 & 1.47E-01 & \textbf{1.09E-01} & 1.10E-01 & 1.47E-01 \\
      &   &   & 7.10E+00 & 7.57E+00 & \textbf{6.02E+00} & 6.33E+00 & \textbf{1.94E-01} & 2.41E-01 & 2.01E-01 & 2.14E-01 \\
      &   &   & 5.67E+00 & 5.37E+00 & \textbf{4.78E+00} & 5.35E+00 & 1.72E-01 & 1.59E-01 & \textbf{1.45E-01} & 1.64E-01 \\ \cline{4-11}
      &   & \multirow{3}{*}{$[-10.0, 10.0]$} & 8.90E+00 & 7.39E+00 & \textbf{6.72E+00} & 8.48E+00 & 3.85E-01 & 2.68E-01 & \textbf{2.61E-01} & 2.89E-01 \\
      &   &   & 1.36E+01 & \textbf{9.32E+00} & 1.08E+01 & 1.15E+01 & 9.89E-01 & \textbf{4.07E-01} & 5.06E-01 & 6.42E-01 \\
      &   &   & 1.14E+01 & \textbf{8.12E+00} & 9.03E+00 & 9.58E+00 & 6.49E-01 & \textbf{3.12E-01} & 3.66E-01 & 3.97E-01 \\
    \midrule
    \multirow{12}{*}{Rag} & \multirow{6}{*}{$7 \times 7$} & \multirow{3}{*}{$[-5.0, 5.0]$} & 3.52E+00 & 3.89E+00 & \textbf{3.06E+00} & 3.96E+00 & 9.64E-02 & 9.47E-02 & \textbf{8.41E-02} & 1.03E-01 \\
      &   &   & \textbf{4.65E+00} & 4.93E+00 & 5.01E+00 & 5.18E+00 & 1.22E-01 & \textbf{1.16E-01} & 1.21E-01 & 1.32E-01 \\
      &   &   & 4.24E+00 & 4.17E+00 & \textbf{3.94E+00} & 4.64E+00 & 1.10E-01 & 1.03E-01 & \textbf{9.90E-02} & 1.20E-01 \\ \cline{4-11}
      &   & \multirow{3}{*}{$[-10.0, 10.0]$} & \textbf{8.12E+00} & 8.18E+00 & 8.24E+00 & 8.78E+00 & 2.17E-01 & \textbf{2.10E-01} & 2.15E-01 & 2.27E-01 \\
      &   &   & 1.01E+01 & 9.24E+00 & \textbf{8.75E+00} & 1.04E+01 & 2.69E-01 & 2.56E-01 & \textbf{2.24E-01} & 2.78E-01 \\
      &   &   & 9.21E+00 & 8.73E+00 & \textbf{8.54E+00} & 9.59E+00 & 2.51E-01 & 2.28E-01 & \textbf{2.20E-01} & 2.61E-01 \\ \cline{4-11}
      & \multirow{6}{*}{$11 \times 11$} & \multirow{3}{*}{$[-5.0, 5.0]$} & 4.97E+00 & \textbf{4.08E+00} & 4.88E+00 & 5.25E+00 & 1.35E-01 & \textbf{1.03E-01} & 1.25E-01 & 1.39E-01 \\
      &   &   & 7.35E+00 & \textbf{5.65E+00} & 6.21E+00 & 6.13E+00 & 2.23E-01 & \textbf{1.51E-01} & 1.69E-01 & 1.54E-01 \\
      &   &   & 6.01E+00 & \textbf{4.93E+00} & 5.41E+00 & 5.56E+00 & 1.68E-01 & \textbf{1.23E-01} & 1.43E-01 & 1.44E-01 \\ \cline{4-11}
      &   & \multirow{3}{*}{$[-10.0, 10.0]$} & 9.58E+00 & 8.43E+00 & \textbf{7.21E+00} & 1.03E+01 & 2.61E-01 & 2.31E-01 & \textbf{2.03E-01} & 2.81E-01 \\
      &   &   & 2.15E+01 & 1.32E+01 & 1.87E+01 & \textbf{1.17E+01} & 8.39E-01 & 3.79E-01 & 5.93E-01 & \textbf{3.54E-01} \\
      &   &   & 1.72E+01 & \textbf{9.74E+00} & 1.07E+01 & 1.10E+01 & 5.99E-01 & \textbf{2.75E-01} & 3.07E-01 & 3.19E-01 \\
    \midrule
    \multirow{12}{*}{Alphabet} & \multirow{6}{*}{$7 \times 7$} & \multirow{3}{*}{$[-5.0, 5.0]$} & 3.81E+00 & \textbf{3.43E+00} & 3.84E+00 & 4.55E+00 & 1.08E-01 & \textbf{9.81E-02} & 1.08E-01 & 1.19E-01 \\
      &   &   & 4.69E+00 & 5.14E+00 & \textbf{4.45E+00} & 5.26E+00 & 1.41E-01 & 1.42E-01 & \textbf{1.23E-01} & 1.67E-01 \\
      &   &   & \textbf{4.14E+00} & 4.15E+00 & 4.21E+00 & 4.90E+00 & 1.21E-01 & 1.23E-01 & \textbf{1.14E-01} & 1.44E-01 \\ \cline{4-11}
      &   & \multirow{3}{*}{$[-10.0, 10.0]$} & 7.46E+00 & \textbf{7.44E+00} & 7.67E+00 & 8.83E+00 & \textbf{2.04E-01} & 2.07E-01 & 2.36E-01 & 2.76E-01 \\
      &   &   & 9.04E+00 & 9.55E+00 & \textbf{8.67E+00} & 1.10E+01 & 2.72E-01 & \textbf{2.47E-01} & 2.76E-01 & 3.45E-01 \\
      &   &   & 8.44E+00 & 8.42E+00 & \textbf{8.33E+00} & 9.93E+00 & 2.36E-01 & \textbf{2.29E-01} & 2.52E-01 & 3.05E-01 \\ \cline{4-11}
      & \multirow{6}{*}{$11 \times 11$} & \multirow{3}{*}{$[-5.0, 5.0]$} & 4.03E+00 & 3.98E+00 & \textbf{3.96E+00} & 5.17E+00 & 1.50E-01 & 1.36E-01 & \textbf{1.34E-01} & 1.95E-01 \\
      &   &   & 5.93E+00 & \textbf{5.17E+00} & 5.21E+00 & 7.07E+00 & 2.42E-01 & 1.74E-01 & \textbf{1.71E-01} & 2.33E-01 \\
      &   &   & 4.86E+00 & 4.76E+00 & \textbf{4.35E+00} & 6.07E+00 & 1.89E-01 & 1.55E-01 & \textbf{1.43E-01} & 2.16E-01 \\ \cline{4-11}
      &   & \multirow{3}{*}{$[-10.0, 10.0]$} & 1.14E+01 & 8.62E+00 & \textbf{7.73E+00} & 9.92E+00 & 3.99E-01 & 3.48E-01 & \textbf{2.15E-01} & 3.71E-01 \\
      &   &   & 1.29E+01 & \textbf{1.12E+01} & 1.15E+01 & 1.28E+01 & 1.01E+00 & 4.63E-01 & \textbf{4.32E-01} & 4.92E-01 \\
      &   &   & 1.19E+01 & \textbf{9.50E+00} & 9.86E+00 & 1.16E+01 & 5.77E-01 & 3.94E-01 & \textbf{3.73E-01} & 4.34E-01 \\
    \midrule
    \multirow{12}{*}{Toad} & \multirow{6}{*}{$7 \times 7$} & \multirow{3}{*}{$[-5.0, 5.0]$} & 1.56E+00 & 1.67E+00 & \textbf{1.22E+00} & 2.17E+00 & \textbf{1.06E-01} & 1.12E-01 & 1.23E-01 & 1.43E-01 \\
      &   &   & 2.58E+00 & 2.40E+00 & \textbf{2.32E+00} & 2.35E+00 & 1.50E-01 & \textbf{1.46E-01} & 1.56E-01 & 1.61E-01 \\
      &   &   & 2.03E+00 & 1.95E+00 & \textbf{1.92E+00} & 2.26E+00 & 1.31E-01 & \textbf{1.25E-01} & 1.35E-01 & 1.51E-01 \\ \cline{4-11}
      &   & \multirow{3}{*}{$[-10.0, 10.0]$} & 3.24E+00 & 2.76E+00 & \textbf{2.62E+00} & 3.53E+00 & 2.85E-01 & 2.73E-01 & \textbf{2.71E-01} & 2.85E-01 \\
      &   &   & 4.97E+00 & 3.78E+00 & \textbf{3.47E+00} & 4.79E+00 & 3.92E-01 & \textbf{3.36E-01} & 3.41E-01 & 3.47E-01 \\
      &   &   & 4.05E+00 & 3.27E+00 & \textbf{3.12E+00} & 4.25E+00 & 3.28E-01 & \textbf{3.05E-01} & 3.14E-01 & 3.18E-01 \\ \cline{4-11}
      & \multirow{6}{*}{$11 \times 11$} & \multirow{3}{*}{$[-5.0, 5.0]$} & 2.19E+00 & 1.90E+00 & \textbf{1.56E+00} & 1.99E+00 & 2.31E-01 & 2.26E-01 & 2.20E-01 & \textbf{2.02E-01} \\
      &   &   & 2.58E+00 & \textbf{2.51E+00} & 2.60E+00 & 3.45E+00 & 3.58E-01 & 2.95E-01 & 3.13E-01 & \textbf{2.90E-01} \\
      &   &   & 2.39E+00 & 2.14E+00 & \textbf{2.10E+00} & 2.67E+00 & 3.21E-01 & 2.68E-01 & \textbf{2.52E-01} & 2.56E-01 \\ \cline{4-11}
      &   & \multirow{3}{*}{$[-10.0, 10.0]$} & 3.06E+00 & 2.97E+00 & \textbf{2.81E+00} & 3.44E+00 & 5.40E-01 & \textbf{4.22E-01} & 4.57E-01 & 4.35E-01 \\
      &   &   & 4.53E+00 & \textbf{3.57E+00} & 4.91E+00 & 4.98E+00 & 8.54E-01 & 7.35E-01 & 7.92E-01 & \textbf{5.01E-01} \\
      &   &   & 3.87E+00 & \textbf{3.23E+00} & 3.73E+00 & 4.08E+00 & 6.72E-01 & 5.18E-01 & 5.77E-01 & \textbf{4.74E-01} \\
    \bottomrule
\end{tabular}
\end{table*}

\begin{figure*}[!t]
\centering
    \includegraphics[width=0.24\linewidth]{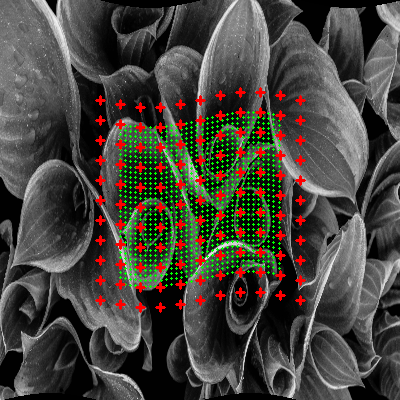}
    \includegraphics[width=0.24\linewidth]{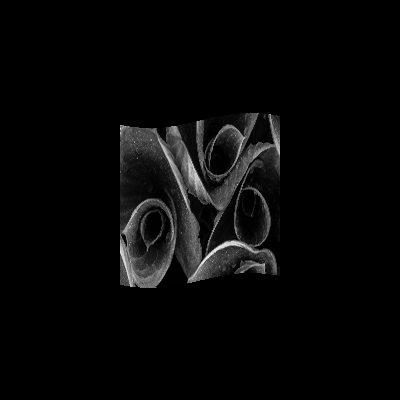}
    \includegraphics[width=0.24\linewidth]{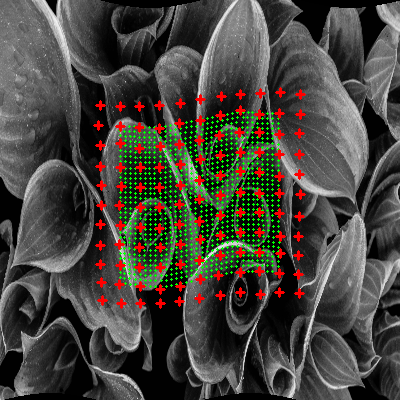}
    \includegraphics[width=0.24\linewidth]{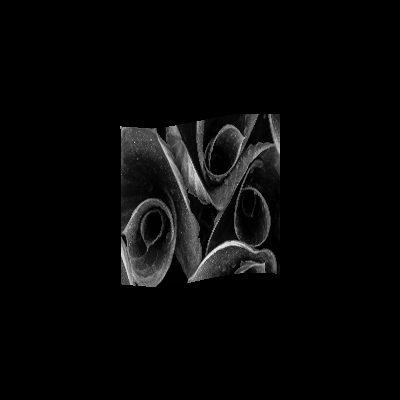} \\ \medskip
    \includegraphics[width=0.24\linewidth]{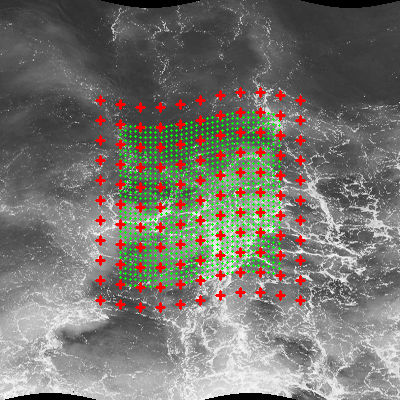}
    \includegraphics[width=0.24\linewidth]{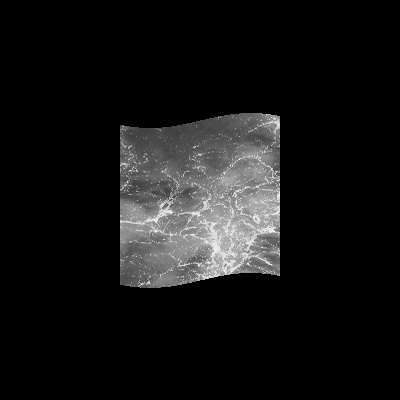}
    \includegraphics[width=0.24\linewidth]{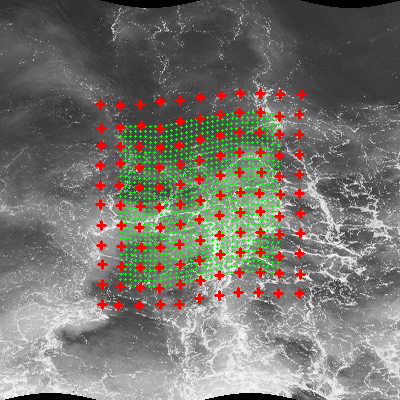}
    \includegraphics[width=0.24\linewidth]{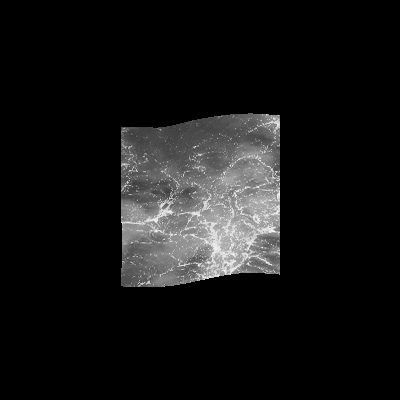} \\ \medskip
    \includegraphics[width=0.24\linewidth]{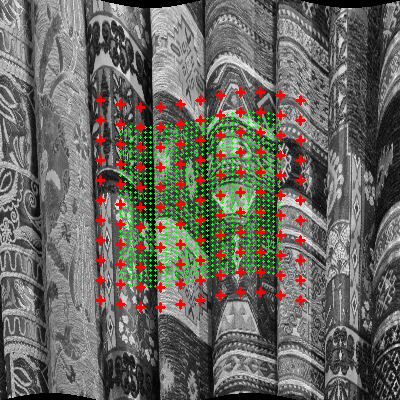}
    \includegraphics[width=0.24\linewidth]{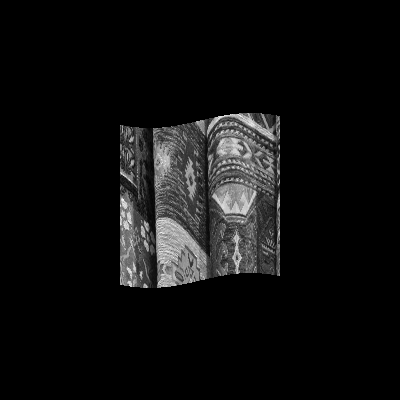}
    \includegraphics[width=0.24\linewidth]{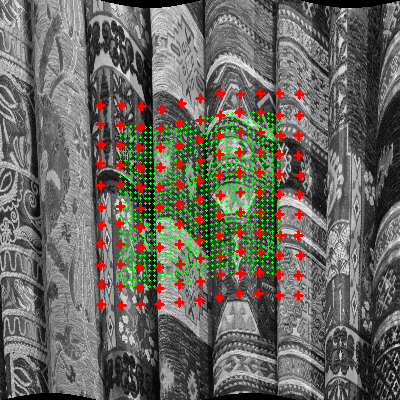}
    \includegraphics[width=0.24\linewidth]{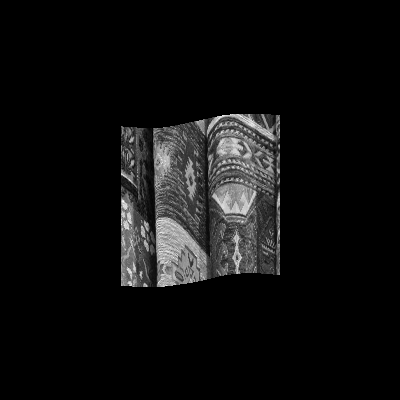} \\ \medskip
    \includegraphics[width=0.24\linewidth]{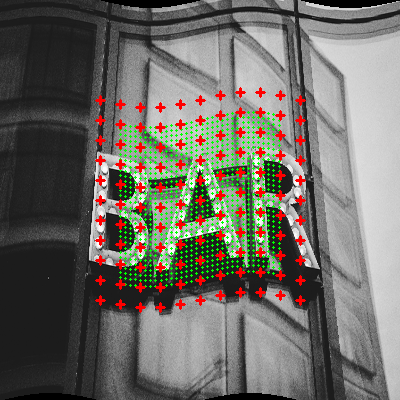}
    \includegraphics[width=0.24\linewidth]{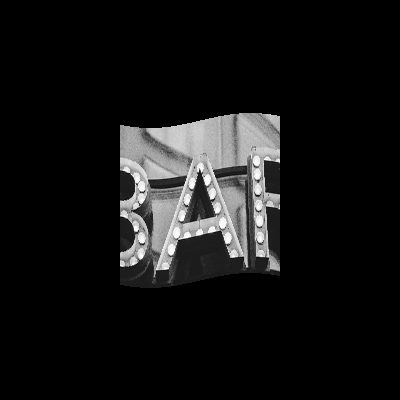}
    \includegraphics[width=0.24\linewidth]{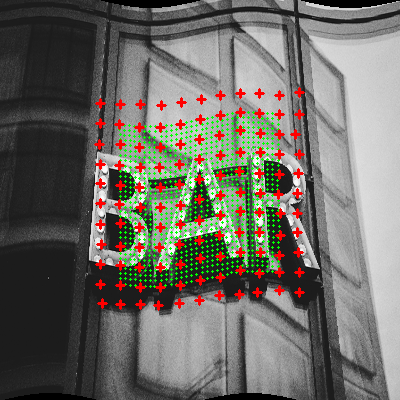}
    \includegraphics[width=0.24\linewidth]{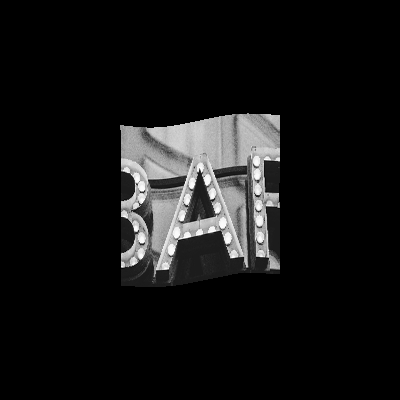} \\ \medskip
    \includegraphics[width=0.24\linewidth]{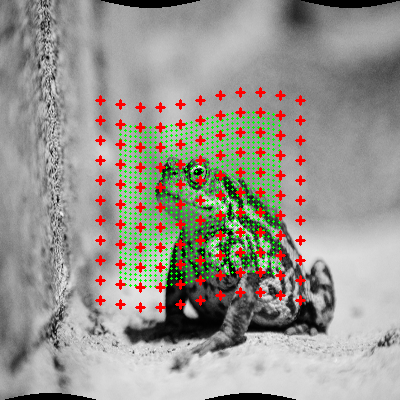}
    \includegraphics[width=0.24\linewidth]{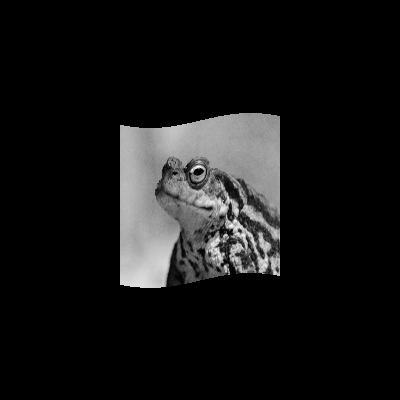}
    \includegraphics[width=0.24\linewidth]{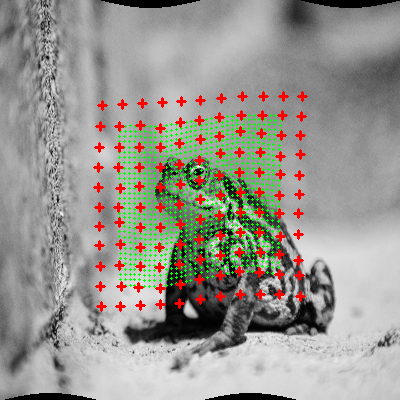}
    \includegraphics[width=0.24\linewidth]{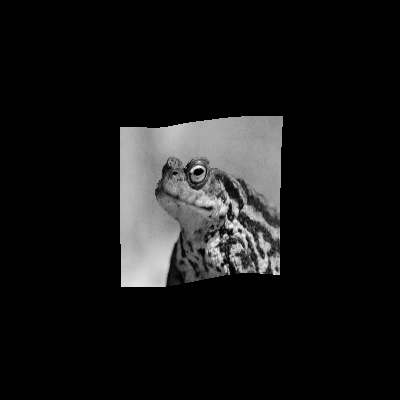}
\caption{Examples of visual results achieved by multi-objective methods on \textit{vertical wavy} images. The 1st$\sim$2nd columns are ground truth and 3rd$\sim$4th columns are results. Red markers represent the control points, and green dots represent the deformed surface generated by forward warping. Images in the 4th column show the deformed template images with estimated deformation.}
\label{fig:vwave_best_success}
\end{figure*}

\subsection{Comparison with Synthetic Data}
\label{sec:experiment_synthetic}
To only focus on verifying the estimation ability rather than robustness against noises, we prepare five images in $400 \times 400$ pixels for generating template images and deformed target images, as shown in Fig. \ref{fig:base_image}. The corresponding template images are obtained by cropping $160 \times 160$ pixels regions from the center of each image. Eight types of deformations are used for generating a deformed target image, the parameters are:
\begin{itemize}
    \item \textbf{Deformation (2 types)}: an image is deformed into a wavy shape by moving the control points according to a sine curve. In addition to vertical-only displacements, a combination of both vertical and horizontal displacements is used.
    \item \textbf{Lattice size (2 types)}: $7 \times 7$ and $11 \times 11$ lattices are used for the bottom image of the pyramid.
    \item \textbf{Ranges of decision variables (2 types)}: we use $[-5.0, 5.0]$ and $[-10.0, 10.0]$ as the range of each decision variable. Ranges are limited to make sure that control points do not overlap with each other spatially.
\end{itemize}
As a result, there are 40 (i.e., 5 images $\times$ 8 types of parameter settings) types of deformed target images in total. These images are generated by using backward warping, and hence the ground truth of displacements can be obtained. In this section, we evaluate each result based on not only the root mean square error (RMSE) but also the mean Euclidean distance error (MEDE). RMSE is calculated from all the pixels between the deformed template image and the sampling region $\omega^\prime$. MEDE is calculated based on all the ground truth displacements.

\begin{figure*}[!t]
\centering
\subfloat[GA]{
    \begin{minipage}{0.49\linewidth}
    \centering
        \includegraphics[width=0.49\hsize]{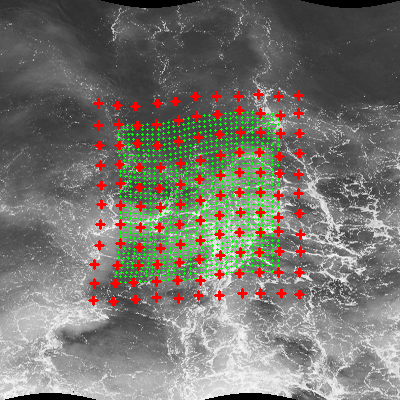}
        \includegraphics[width=0.49\hsize]{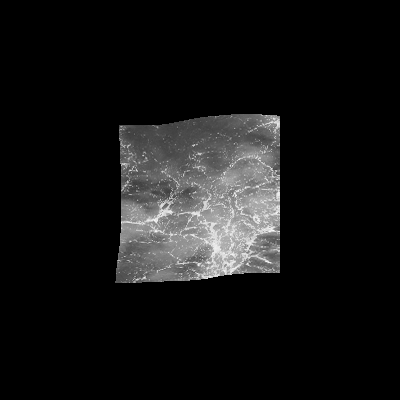}
    \end{minipage}
    \label{fig:vwave_best_sea_ga}
}
\subfloat[two-objective NSGA-II]{
    \begin{minipage}{0.49\linewidth}
    \centering
        \includegraphics[width=0.49\hsize]{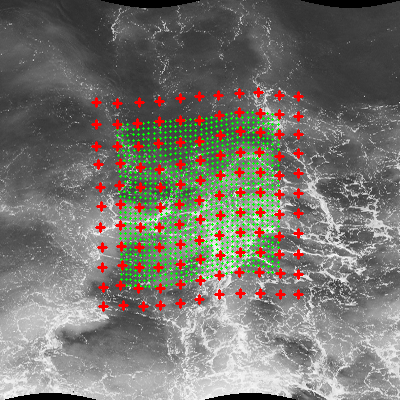}
        \includegraphics[width=0.49\hsize]{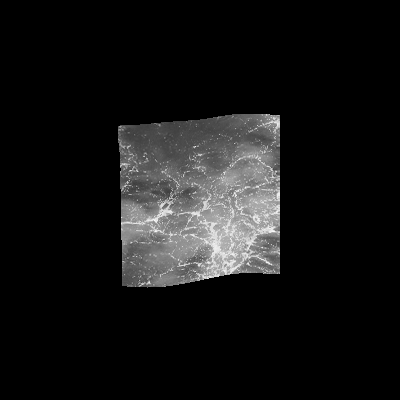}
    \end{minipage}
    \label{fig:vwave_best_sea_nsgaii}
}\\
\subfloat[two-objective NSGA-III]{
    \begin{minipage}{0.49\linewidth}
    \centering
        \includegraphics[width=0.49\hsize]{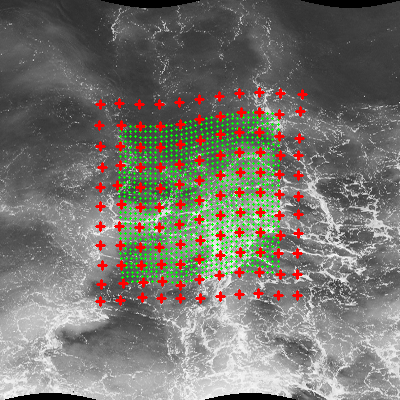}
        \includegraphics[width=0.49\hsize]{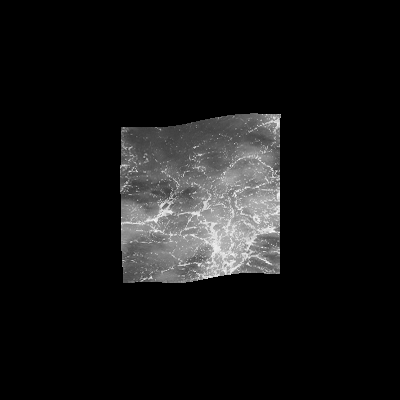}
    \end{minipage}
    \label{fig:vwave_best_sea_nsgaiii_2objs}
}
\subfloat[four-objective NSGA-III]{
    \begin{minipage}{0.49\linewidth}
    \centering
        \includegraphics[width=0.49\hsize]{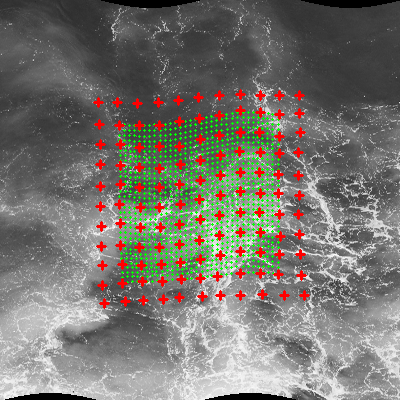}
        \includegraphics[width=0.49\hsize]{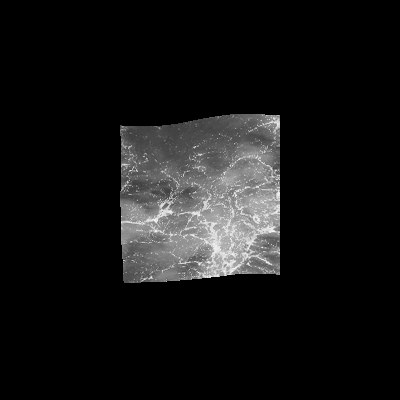}
    \end{minipage}
    \label{fig:vwave_best_sea_nsgaiii_4objs}
}
\caption{Comparison of the results showing maximum MEDE for the sea image with the $11 \times 11$ lattice and $[-10.0, 10.0]$ range.}
\label{fig:vwave_best_sea}
\end{figure*}

\subsubsection{Results of Vertical Wavy Images}
\label{sec:experiment_synthetic_vwave}
Comprehensive numerical results of the best solutions with respect to the vertical wavy images are summarized in Table \ref{tab:vwave_best}. For each combination of image setting and algorithm, the minimum, maximum, and average values based on five random trials are investigated. As can be observed by comparing the four algorithms in terms of RMSE and MEDE, it is clear that the two-objective algorithms can achieve better results in most cases. Examples of visual results are shown in Fig \ref{fig:vwave_best_success}, from which we can observe that these methods can estimate deformation parameters correctly for all the test images. By contrast, GA with single-objective achieves the best result only once in terms of the average RMSE and zero times in terms of the average MEDE value. The accuracy of GA can degrade significantly, e.g., $11 \times 11$ lattice with $[-10.0, 10.0]$ range as illustrated in Fig. \ref{fig:vwave_best_sea_ga}. These observations verify the effectiveness of the multi-objective approaches. In addition, we can observe that four-objective NSGA-III gets poorer results than two-objective algorithms. Focusing on average values of both evaluation criteria, four-objective NSGA-III outperforms others for zero times regarding the RMSE and only once regarding the MEDE.

The final solutions after post-processing on Pareto optimal solutions obtained by multi-objective algorithms are compared in Table \ref{tab:vwave_compounded}. As can be observed, the post-processing successfully improves the estimation accuracy in many cases comparing to Table \ref{tab:vwave_best}. In particular, four-objective NSGA-III benefits most from the post-processing. Focusing on the number of improved results on the average value, four-objective NSGA-III performs the best 17 times on RMSE and 19 times on MEDE, while two-objective NSGA-II performs the best for 15 and 14 times, respectively.

\begin{table*}[!t]
\centering
\caption{Comparative results of the final solutions after post-processing with respect to the \textit{vertical wavy deformation}. Each cell consists of three values which represent the minimum, maximum, and average values based on five random trials from top to bottom. The values smaller than the corresponding values in Table \ref{tab:vwave_best} with the same settings are shown in Italics. The minimum value in terms of RMSE and MEDE with respect to each row, including the results in Table \ref{tab:vwave_best}, is emphasized in bold.}
\label{tab:vwave_compounded}
\scriptsize
    \begin{tabular}{cccrrrrrr}
    \toprule
      &   &   & \multicolumn{3}{c}{RMSE} & \multicolumn{3}{c}{MEDE} \\
    \multirow{3}{*}{Image} & \multirow{3}{*}{\begin{tabular}{c} Lattice \\ size \end{tabular}} & Decision & \multicolumn{1}{c}{Two} & \multicolumn{1}{c}{Two} & \multicolumn{1}{c}{Four} & \multicolumn{1}{c}{Two} & \multicolumn{1}{c}{Two} & \multicolumn{1}{c}{Four} \\
      &   & variable & \multicolumn{1}{c}{objective} & \multicolumn{1}{c}{objective} & \multicolumn{1}{c}{objective} & \multicolumn{1}{c}{objective} & \multicolumn{1}{c}{objective} & \multicolumn{1}{c}{objective} \\
      &   & range & \multicolumn{1}{c}{NSGA-II} & \multicolumn{1}{c}{NSGA-III} & \multicolumn{1}{c}{NSGA-III} & \multicolumn{1}{c}{NSGA-II} & \multicolumn{1}{c}{NSGA-III} & \multicolumn{1}{c}{NSGA-III} \\
    \midrule
    \midrule
    \multirow{12}{*}{Plant} & \multirow{6}{*}{$7 \times 7$} & \multirow{3}{*}{$[-5.0, 5.0]$} & \textit{\textbf{1.98E+00}} & 2.40E+00 & \textit{2.37E+00} & \textit{\textbf{9.98E-02}} & 1.08E-01 & \textit{1.16E-01} \\
      &   &   & 3.04E+00 & 2.85E+00 & 3.40E+00 & 1.31E-01 & 1.38E-01 & \textit{1.75E-01} \\
      &   &   & 2.45E+00 & 2.54E+00 & \textit{2.80E+00} & 1.15E-01 & 1.20E-01 & \textit{1.37E-01} \\ \cline{4-9}
      &   & \multirow{3}{*}{$[-10.0, 10.0]$} & 4.50E+00 & \textit{\textbf{4.29E+00}} & \textit{4.98E+00} & 2.28E-01 & \textit{2.23E-01} & \textit{2.59E-01} \\
      &   &   & 5.61E+00 & 6.71E+00 & \textit{6.81E+00} & \textit{\textbf{2.99E-01}} & 3.42E-01 & \textit{3.45E-01} \\
      &   &   & 5.12E+00 & \textit{5.15E+00} & \textit{5.70E+00} & 2.67E-01 & 2.70E-01 & \textit{3.01E-01} \\ \cline{4-9}
      & \multirow{6}{*}{$11 \times 11$} & \multirow{3}{*}{$[-5.0, 5.0]$} & \textit{2.55E+00} & \textit{\textbf{2.25E+00}} & \textit{3.07E+00} & \textit{1.37E-01} & \textit{\textbf{1.04E-01}} & \textit{1.58E-01} \\
      &   &   & 4.27E+00 & \textit{\textbf{3.51E+00}} & \textit{3.79E+00} & 2.16E-01 & \textit{\textbf{1.85E-01}} & \textit{1.96E-01} \\
      &   &   & \textit{3.23E+00} & \textit{\textbf{3.03E+00}} & \textit{3.38E+00} & \textit{1.76E-01} & \textit{\textbf{1.57E-01}} & \textit{1.74E-01} \\ \cline{4-9}
      &   & \multirow{3}{*}{$[-10.0, 10.0]$} & 5.21E+00 & 5.27E+00 & \textit{6.04E+00} & 2.86E-01 & 3.48E-01 & \textit{3.25E-01} \\
      &   &   & \textit{9.51E+00} & \textit{\textbf{7.49E+00}} & \textit{8.34E+00} & 7.39E-01 & 5.27E-01 & \textit{\textbf{5.03E-01}} \\
      &   &   & \textit{6.73E+00} & \textit{\textbf{6.61E+00}} & \textit{6.87E+00} & \textit{4.53E-01} & \textit{4.29E-01} & \textit{\textbf{4.26E-01}} \\
    \midrule
    \multirow{12}{*}{Sea} & \multirow{6}{*}{$7 \times 7$} & \multirow{3}{*}{$[-5.0, 5.0]$} & \textit{3.36E+00} & \textit{\textbf{3.19E+00}} & \textit{4.11E+00} & \textit{9.53E-02} & 1.00E-01 & \textit{1.26E-01} \\
      &   &   & \textit{\textbf{3.78E+00}} & \textit{4.18E+00} & 5.00E+00 & 1.23E-01 & \textit{1.21E-01} & \textit{1.54E-01} \\
      &   &   & \textit{\textbf{3.61E+00}} & \textit{3.71E+00} & 4.53E+00 & \textit{1.11E-01} & \textit{\textbf{1.10E-01}} & \textit{1.41E-01} \\ \cline{4-9}
      &   & \multirow{3}{*}{$[-10.0, 10.0]$} & 7.73E+00 & 8.15E+00 & \textit{8.45E+00} & 2.34E-01 & \textit{2.43E-01} & \textit{2.55E-01} \\
      &   &   & 9.29E+00 & \textit{9.69E+00} & 1.11E+01 & 2.67E-01 & \textit{3.40E-01} & 3.75E-01 \\
      &   &   & \textit{\textbf{8.32E+00}} & \textit{8.94E+00} & \textit{9.45E+00} & 2.46E-01 & 3.04E-01 & \textit{3.09E-01} \\ \cline{4-9}
      & \multirow{6}{*}{$11 \times 11$} & \multirow{3}{*}{$[-5.0, 5.0]$} & \textit{4.00E+00} & \textit{\textbf{3.69E+00}} & \textit{4.46E+00} & 1.09E-01 & \textit{\textbf{1.03E-01}} & \textit{1.27E-01} \\
      &   &   & \textit{7.43E+00} & \textit{5.68E+00} & \textit{\textbf{5.59E+00}} & \textit{2.34E-01} & \textit{1.86E-01} & \textit{\textbf{1.85E-01}} \\
      &   &   & \textit{5.21E+00} & \textit{\textbf{4.58E+00}} & \textit{4.92E+00} & \textit{1.56E-01} & \textit{\textbf{1.39E-01}} & \textit{1.43E-01} \\ \cline{4-9}
      &   & \multirow{3}{*}{$[-10.0, 10.0]$} & \textit{6.72E+00} & \textit{\textbf{6.38E+00}} & \textit{7.84E+00} & \textit{2.45E-01} & \textit{\textbf{2.42E-01}} & \textit{2.55E-01} \\
      &   &   & 9.33E+00 & \textit{1.07E+01} & \textit{1.11E+01} & \textit{\textbf{4.05E-01}} & \textit{5.03E-01} & \textit{5.94E-01} \\
      &   &   & \textit{\textbf{7.83E+00}} & \textit{8.90E+00} & \textit{9.12E+00} & \textit{\textbf{3.00E-01}} & \textit{3.54E-01} & \textit{3.60E-01} \\
    \midrule
    \multirow{12}{*}{Rag} & \multirow{6}{*}{$7 \times 7$} & \multirow{3}{*}{$[-5.0, 5.0]$} & 3.90E+00 & 3.17E+00 & 4.08E+00 & \textit{9.39E-02} & 8.83E-02 & \textit{1.03E-01} \\
      &   &   & 5.02E+00 & \textit{4.68E+00} & \textit{5.18E+00} & 1.29E-01 & \textit{1.16E-01} & 1.44E-01 \\
      &   &   & 4.25E+00 & \textit{\textbf{3.77E+00}} & 4.67E+00 & 1.08E-01 & \textit{\textbf{9.79E-02}} & 1.21E-01 \\ \cline{4-9}
      &   & \multirow{3}{*}{$[-10.0, 10.0]$} & \textit{\textbf{7.59E+00}} & 8.27E+00 & 8.78E+00 & \textit{\textbf{2.08E-01}} & 2.17E-01 & 2.33E-01 \\
      &   &   & 9.32E+00 & 9.16E+00 & \textit{9.85E+00} & 2.58E-01 & 2.38E-01 & \textit{2.57E-01} \\
      &   &   & \textit{\textbf{8.44E+00}} & 8.65E+00 & \textit{9.45E+00} & \textit{2.22E-01} & 2.27E-01 & \textit{2.49E-01} \\ \cline{4-9}
      & \multirow{6}{*}{$11 \times 11$} & \multirow{3}{*}{$[-5.0, 5.0]$} & \textit{\textbf{4.04E+00}} & \textit{4.86E+00} & \textit{4.72E+00} & \textit{\textbf{1.02E-01}} & \textit{1.21E-01} & \textit{1.24E-01} \\
      &   &   & \textit{\textbf{5.47E+00}} & \textit{5.95E+00} & \textit{5.82E+00} & \textit{1.43E-01} & \textit{1.62E-01} & \textit{\textbf{1.41E-01}} \\
      &   &   & \textit{\textbf{4.70E+00}} & \textit{5.33E+00} & \textit{5.23E+00} & \textit{\textbf{1.18E-01}} & \textit{1.38E-01} & \textit{1.31E-01} \\ \cline{4-9}
      &   & \multirow{3}{*}{$[-10.0, 10.0]$} & \textit{8.14E+00} & \textit{\textbf{6.86E+00}} & \textit{9.25E+00} & 2.33E-01 & \textit{\textbf{1.95E-01}} & \textit{2.52E-01} \\
      &   &   & 1.35E+01 & \textit{1.86E+01} & \textit{\textbf{1.14E+01}} & 3.93E-01 & \textit{5.86E-01} & \textit{\textbf{3.42E-01}} \\
      &   &   & \textit{\textbf{9.66E+00}} & \textit{1.03E+01} & \textit{1.02E+01} & \textit{\textbf{2.73E-01}} & \textit{2.99E-01} & \textit{2.89E-01} \\
    \midrule
    \multirow{12}{*}{Alphabet} & \multirow{6}{*}{$7 \times 7$} & \multirow{3}{*}{$[-5.0, 5.0]$} & 3.58E+00 & \textit{3.57E+00} & \textit{4.30E+00} & 1.08E-01 & \textit{1.07E-01} & \textit{1.18E-01} \\
      &   &   & \textit{4.86E+00} & 4.49E+00 & \textit{5.02E+00} & \textit{1.34E-01} & 1.29E-01 & \textit{1.55E-01} \\
      &   &   & \textit{\textbf{4.11E+00}} & \textit{4.14E+00} & \textit{4.72E+00} & \textit{1.21E-01} & 1.20E-01 & \textit{1.40E-01} \\ \cline{4-9}
      &   & \multirow{3}{*}{$[-10.0, 10.0]$} & \textit{\textbf{7.28E+00}} & 7.85E+00 & 8.84E+00 & \textit{\textbf{1.93E-01}} & \textit{2.35E-01} & \textit{2.57E-01} \\
      &   &   & \textit{9.51E+00} & \textit{\textbf{8.62E+00}} & \textit{1.02E+01} & 2.73E-01 & 2.90E-01 & \textit{3.24E-01} \\
      &   &   & \textit{\textbf{8.10E+00}} & \textit{8.22E+00} & \textit{9.58E+00} & 2.31E-01 & 2.54E-01 & \textit{2.91E-01} \\ \cline{4-9}
      & \multirow{6}{*}{$11 \times 11$} & \multirow{3}{*}{$[-5.0, 5.0]$} & 4.00E+00 & \textit{\textbf{3.86E+00}} & \textit{5.01E+00} & \textit{1.34E-01} & \textit{\textbf{1.31E-01}} & \textit{1.79E-01} \\
      &   &   & \textit{5.04E+00} & \textit{\textbf{4.95E+00}} & \textit{6.51E+00} & 1.81E-01 & \textit{\textbf{1.61E-01}} & \textit{2.04E-01} \\
      &   &   & \textit{4.64E+00} & \textit{\textbf{4.26E+00}} & \textit{5.73E+00} & \textit{1.52E-01} & \textit{\textbf{1.39E-01}} & \textit{1.93E-01} \\ \cline{4-9}
      &   & \multirow{3}{*}{$[-10.0, 10.0]$} & \textit{8.45E+00} & \textit{\textbf{7.47E+00}} & \textit{9.00E+00} & 3.51E-01 & \textit{\textbf{2.14E-01}} & \textit{3.46E-01} \\
      &   &   & \textit{\textbf{1.10E+01}} & \textit{1.13E+01} & \textit{1.20E+01} & \textit{4.50E-01} & \textit{\textbf{4.22E-01}} & \textit{4.66E-01} \\
      &   &   & 9.56E+00 & \textit{9.68E+00} & \textit{1.10E+01} & \textit{3.88E-01} & \textit{\textbf{3.66E-01}} & \textit{4.02E-01} \\
    \midrule
    \multirow{12}{*}{Toad} & \multirow{6}{*}{$7 \times 7$} & \multirow{3}{*}{$[-5.0, 5.0]$} & \textit{1.55E+00} & \textit{\textbf{1.21E+00}} & \textit{1.91E+00} & \textit{\textbf{1.01E-01}} & \textit{1.11E-01} & \textit{1.25E-01} \\
      &   &   & 2.52E+00 & 2.39E+00 & 2.58E+00 & \textit{\textbf{1.38E-01}} & \textit{1.49E-01} & \textit{1.55E-01} \\
      &   &   & \textit{1.93E+00} & \textit{\textbf{1.91E+00}} & \textit{2.24E+00} & \textit{\textbf{1.21E-01}} & \textit{1.31E-01} & \textit{1.39E-01} \\ \cline{4-9}
      &   & \multirow{3}{*}{$[-10.0, 10.0]$} & 2.99E+00 & \textit{\textbf{2.44E+00}} & 3.78E+00 & \textit{2.56E-01} & 2.79E-01 & \textit{\textbf{2.54E-01}} \\
      &   &   & 4.29E+00 & 3.62E+00 & 5.14E+00 & \textit{\textbf{3.35E-01}} & 3.52E-01 & 3.52E-01 \\
      &   &   & 3.68E+00 & \textit{\textbf{3.12E+00}} & 4.64E+00 & \textit{\textbf{2.91E-01}} & 3.14E-01 & \textit{3.03E-01} \\ \cline{4-9}
      & \multirow{6}{*}{$11 \times 11$} & \multirow{3}{*}{$[-5.0, 5.0]$} & \textit{1.86E+00} & \textit{\textbf{1.51E+00}} & \textit{1.90E+00} & 2.33E-01 & \textit{2.08E-01} & \textit{\textbf{1.76E-01}} \\
      &   &   & \textit{\textbf{2.50E+00}} & 2.64E+00 & \textit{3.40E+00} & \textit{2.88E-01} & \textit{3.04E-01} & \textit{\textbf{2.80E-01}} \\
      &   &   & \textit{2.12E+00} & 2.13E+00 & \textit{2.60E+00} & \textit{2.65E-01} & \textit{2.45E-01} & \textit{\textbf{2.36E-01}} \\ \cline{4-9}
      &   & \multirow{3}{*}{$[-10.0, 10.0]$} & \textit{2.92E+00} & \textit{\textbf{2.79E+00}} & \textit{3.18E+00} & 4.26E-01 & 4.71E-01 & \textit{\textbf{4.05E-01}} \\
      &   &   & 3.67E+00 & \textit{4.72E+00} & \textit{4.81E+00} & 7.59E-01 & \textit{7.87E-01} & \textit{\textbf{4.80E-01}} \\
      &   &   & \textit{\textbf{3.20E+00}} & \textit{3.64E+00} & \textit{3.94E+00} & 5.27E-01 & \textit{5.70E-01} & \textit{\textbf{4.45E-01}} \\
    \bottomrule
    \end{tabular}
\end{table*}

\subsubsection{Results of Vertical and Horizontal Wavy Images}
\label{sec:experiment_synthetic_vhwave}
\begin{figure*}[!t]
\centering
    \includegraphics[width=0.24\linewidth]{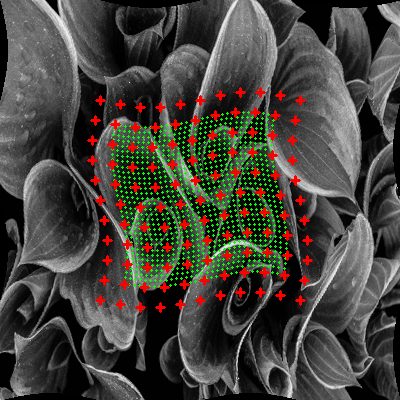}
    \includegraphics[width=0.24\linewidth]{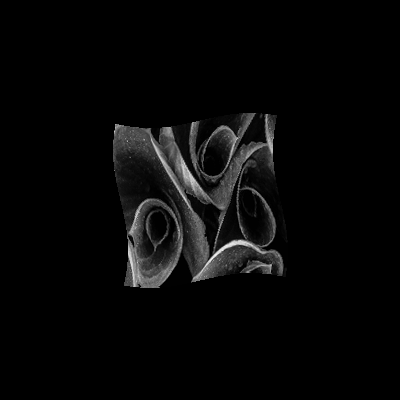}
    \includegraphics[width=0.24\linewidth]{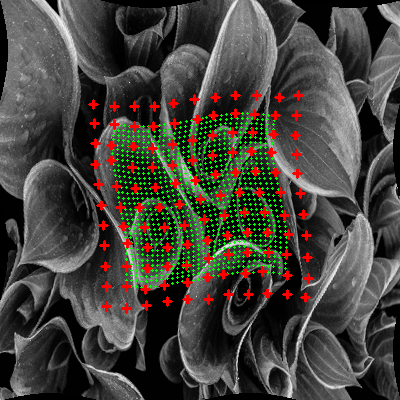}
    \includegraphics[width=0.24\linewidth]{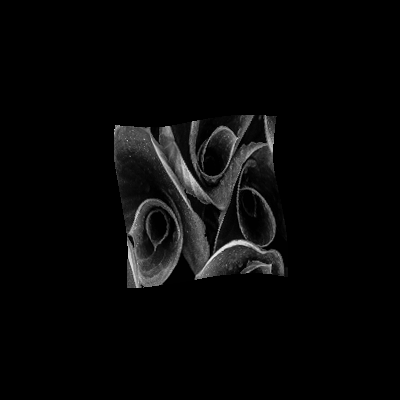} \\ \medskip
    \includegraphics[width=0.24\linewidth]{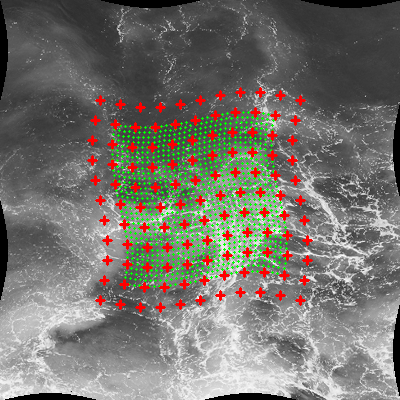}
    \includegraphics[width=0.24\linewidth]{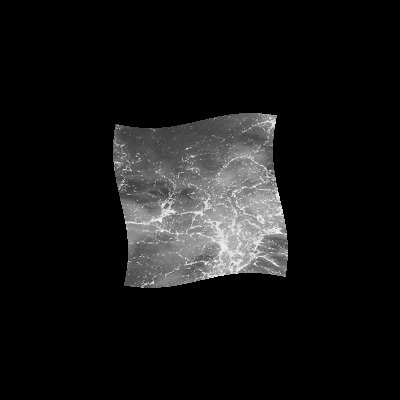}
    \includegraphics[width=0.24\linewidth]{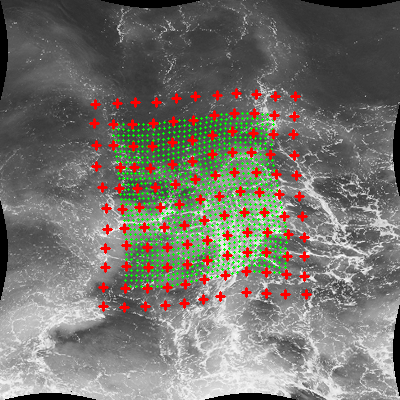}
    \includegraphics[width=0.24\linewidth]{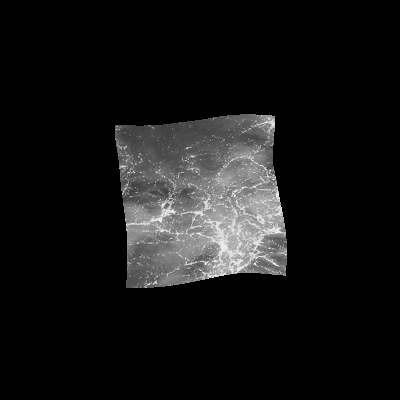} \\ \medskip
    \includegraphics[width=0.24\linewidth]{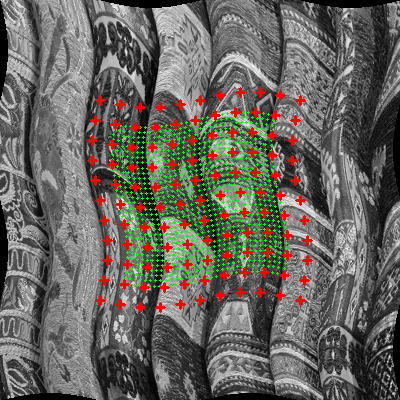}
    \includegraphics[width=0.24\linewidth]{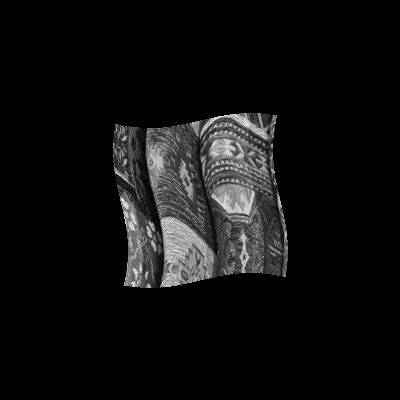}
    \includegraphics[width=0.24\linewidth]{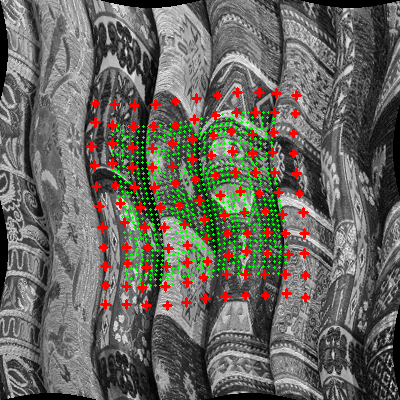}
    \includegraphics[width=0.24\linewidth]{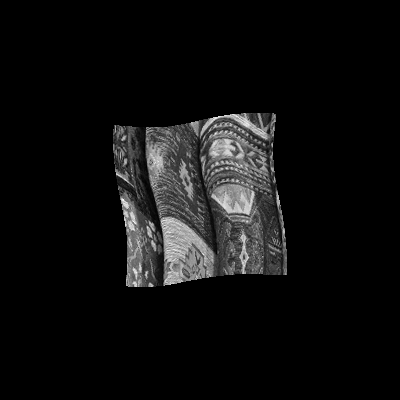} \\ \medskip
    \includegraphics[width=0.24\linewidth]{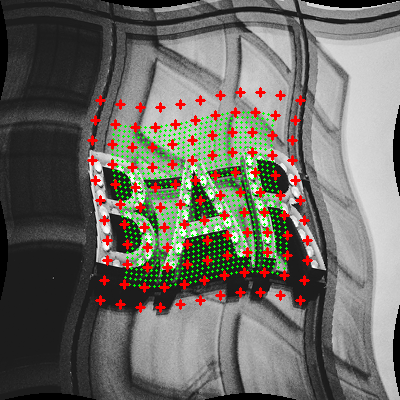}
    \includegraphics[width=0.24\linewidth]{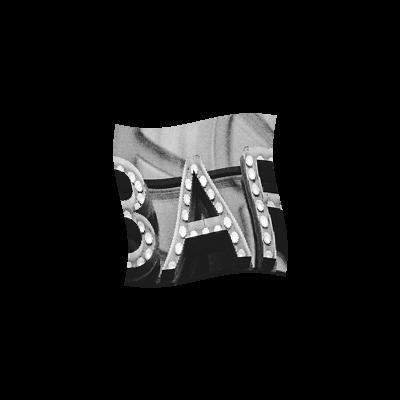}
    \includegraphics[width=0.24\linewidth]{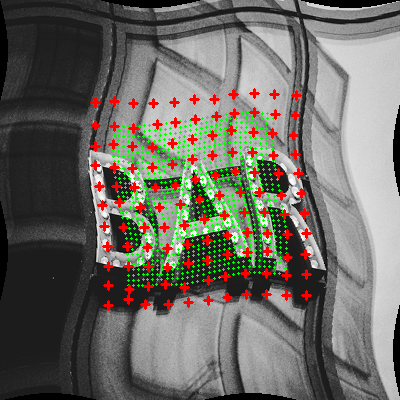}
    \includegraphics[width=0.24\linewidth]{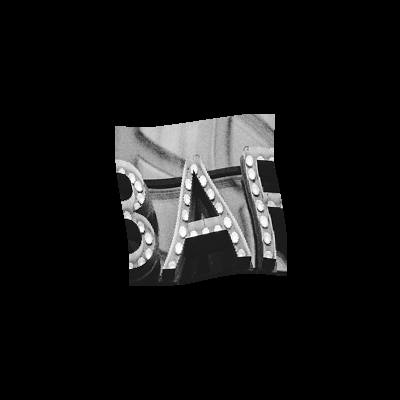} \\ \medskip
    \includegraphics[width=0.24\linewidth]{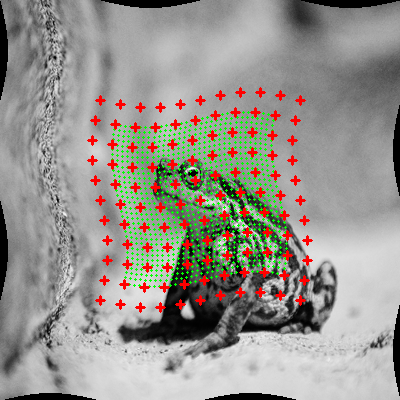}
    \includegraphics[width=0.24\linewidth]{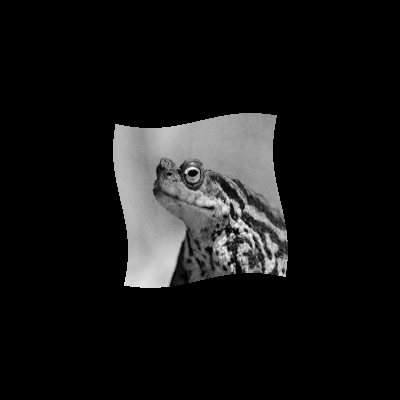}
    \includegraphics[width=0.24\linewidth]{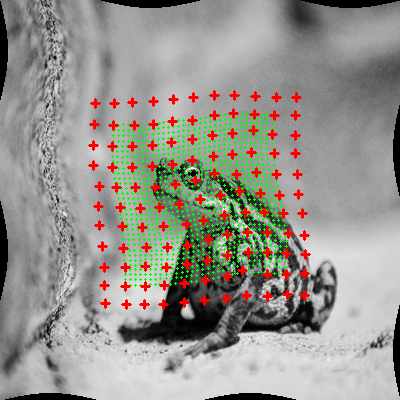}
    \includegraphics[width=0.24\linewidth]{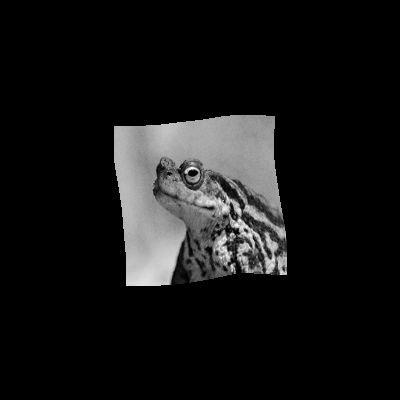}
\caption{Examples of visual results achieved by multi-objective methods on \textit{horizontal and vertical wavy} images. The 1st$\sim$2nd columns are ground truth and 3rd$\sim$4th columns are results. Red markers represent the control points, and green dots represent the deformed surface generated by forward warping. Images in the 4th column show the deformed template images with estimated deformation.}
\label{fig:vhwave_best_success}
\end{figure*}

The results of the best solutions with respect to the vertical and horizontal wavy images are shown in Table \ref{tab:vhwave_best}. Despite more complex deformations, the multi-objective approaches can still achieve good estimation. Several qualitative results are shown in Fig. \ref{fig:vhwave_best_success}. Comparing with Table \ref{tab:vwave_best}, we can observe that the best results are irregularly distributed in terms of both evaluation criteria. Nevertheless, two-objective NSGA-II and NSGA-III are still better choices overall. In the case of $7 \times 7$ lattice, GA can achieve the best result when the search space is small (e.g., the sea image). Four-objective NSGA-III shows good results in the case of $11 \times 11$ lattice setting, which achieves the best average MEDE for six times out of ten times. Although the number of groups is a hyper-parameter to be handled carefully as mentioned in Sec. \ref{sec:experiment_synthetic_vwave}, our results show the trend that larger number of groups is more effective when dealing with complex and subtle deformations.

\begin{table*}[!t]
\centering
\caption{Comparative results of the best solutions with respect to the \textit{vertical and horizontal wavy deformation}. Each cell consists of three values which represent the minimum, maximum, and average values based on five random trials from top to bottom. The best results in terms of RMSE and MEDE with respect to each row are emphasized in bold.}
\label{tab:vhwave_best}
\scriptsize
    \begin{tabular}{cccrrrrrrrr}
    \toprule
      &   &   & \multicolumn{4}{c}{RMSE} & \multicolumn{4}{c}{MEDE} \\
    \multirow{3}{*}{Image} & \multirow{3}{*}{\begin{tabular}{c} Lattice \\ size \end{tabular}} & Decision & \multicolumn{1}{c}{\multirow{3}{*}{GA}} & \multicolumn{1}{c}{Two} & \multicolumn{1}{c}{Two} & \multicolumn{1}{c}{Four} & \multicolumn{1}{c}{\multirow{3}{*}{GA}} & \multicolumn{1}{c}{Two} & \multicolumn{1}{c}{Two} & \multicolumn{1}{c}{Four} \\
      &   & variable &   & \multicolumn{1}{c}{objective} & \multicolumn{1}{c}{objective} & \multicolumn{1}{c}{objective} &   & \multicolumn{1}{c}{objective} & \multicolumn{1}{c}{objective} & \multicolumn{1}{c}{objective} \\
      &   & range &   & \multicolumn{1}{c}{NSGA-II} & \multicolumn{1}{c}{NSGA-III} & \multicolumn{1}{c}{NSGA-III} &   & \multicolumn{1}{c}{NSGA-II} & \multicolumn{1}{c}{NSGA-III} & \multicolumn{1}{c}{NSGA-III} \\
    \midrule
    \midrule
    \multirow{12}{*}{Plant} & \multirow{6}{*}{$7 \times 7$} & \multirow{3}{*}{$[-5.0, 5.0]$} & 5.13E+00 & \textbf{4.38E+00} & 4.49E+00 & 4.95E+00 & 2.29E-01 & \textbf{1.89E-01} & 2.06E-01 & 2.12E-01 \\
      &   &   & \textbf{5.58E+00} & 6.89E+00 & 5.65E+00 & 7.17E+00 & 2.64E-01 & 3.45E-01 & \textbf{2.55E-01} & 3.27E-01 \\
      &   &   & 5.39E+00 & 5.45E+00 & \textbf{5.01E+00} & 5.80E+00 & 2.44E-01 & 2.44E-01 & \textbf{2.30E-01} & 2.66E-01 \\ \cline{4-11}
      &   & \multirow{3}{*}{$[-10.0, 10.0]$} & \textbf{9.42E+00} & 9.56E+00 & 9.51E+00 & 1.11E+01 & 4.94E-01 & 5.27E-01 & \textbf{4.86E-01} & 6.19E-01 \\
      &   &   & \textbf{1.05E+01} & 1.08E+01 & 1.16E+01 & 1.25E+01 & 6.06E-01 & \textbf{5.94E-01} & 6.96E-01 & 6.90E-01 \\
      &   &   & \textbf{1.01E+01} & 1.01E+01 & 1.05E+01 & 1.16E+01 & \textbf{5.53E-01} & 5.66E-01 & 5.71E-01 & 6.57E-01 \\ \cline{4-11}
      & \multirow{6}{*}{$11 \times 11$} & \multirow{3}{*}{$[-5.0, 5.0]$} & 8.40E+00 & 6.10E+00 & 6.28E+00 & \textbf{5.13E+00} & 5.13E-01 & 3.32E-01 & 3.57E-01 & \textbf{2.55E-01} \\
      &   &   & 9.95E+00 & 8.99E+00 & 8.92E+00 & \textbf{8.79E+00} & 5.76E-01 & 5.43E-01 & \textbf{5.18E-01} & 5.27E-01 \\
      &   &   & 8.95E+00 & 7.82E+00 & 7.50E+00 & \textbf{7.37E+00} & 5.41E-01 & 4.48E-01 & 4.36E-01 & \textbf{4.14E-01} \\ \cline{4-11}
      &   & \multirow{3}{*}{$[-10.0, 10.0]$} & 1.02E+01 & \textbf{8.32E+00} & 9.84E+00 & 9.94E+00 & 6.07E-01 & \textbf{5.18E-01} & 6.63E-01 & 6.22E-01 \\
      &   &   & 2.28E+01 & 1.22E+01 & 1.34E+01 & \textbf{1.17E+01} & 3.00E+00 & 7.95E-01 & 9.62E-01 & \textbf{7.07E-01} \\
      &   &   & 1.36E+01 & \textbf{9.97E+00} & 1.14E+01 & 1.05E+01 & 1.20E+00 & \textbf{6.44E-01} & 7.45E-01 & 6.80E-01 \\
    \midrule
    \multirow{12}{*}{Sea} & \multirow{6}{*}{$7 \times 7$} & \multirow{3}{*}{$[-5.0, 5.0]$} & \textbf{5.32E+00} & 5.41E+00 & 5.65E+00 & 7.00E+00 & \textbf{1.60E-01} & 1.76E-01 & 1.62E-01 & 2.06E-01 \\
      &   &   & \textbf{6.29E+00} & 6.78E+00 & 6.40E+00 & 8.50E+00 & \textbf{1.98E-01} & 2.21E-01 & 2.15E-01 & 3.12E-01 \\
      &   &   & \textbf{5.78E+00} & 6.15E+00 & 6.07E+00 & 7.54E+00 & \textbf{1.72E-01} & 1.96E-01 & 1.97E-01 & 2.55E-01 \\ \cline{4-11}
      &   & \multirow{3}{*}{$[-10.0, 10.0]$} & \textbf{1.09E+01} & 1.14E+01 & 1.23E+01 & 1.26E+01 & 4.94E-01 & \textbf{4.90E-01} & 5.35E-01 & 5.90E-01 \\
      &   &   & 1.36E+01 & \textbf{1.32E+01} & 1.46E+01 & 1.45E+01 & 6.47E-01 & \textbf{5.77E-01} & 9.34E-01 & 8.52E-01 \\
      &   &   & 1.25E+01 & \textbf{1.21E+01} & 1.34E+01 & 1.32E+01 & 5.78E-01 & \textbf{5.28E-01} & 6.60E-01 & 6.60E-01 \\ \cline{4-11}
      & \multirow{6}{*}{$11 \times 11$} & \multirow{3}{*}{$[-5.0, 5.0]$} & 6.38E+00 & 6.85E+00 & \textbf{5.99E+00} & 7.31E+00 & 2.73E-01 & 2.81E-01 & \textbf{2.43E-01} & 2.70E-01 \\
      &   &   & 9.62E+00 & \textbf{9.61E+00} & 9.90E+00 & 1.00E+01 & 4.97E-01 & 4.65E-01 & 5.17E-01 & \textbf{4.62E-01} \\
      &   &   & \textbf{8.07E+00} & 8.41E+00 & 8.82E+00 & 8.47E+00 & 3.73E-01 & 3.90E-01 & 4.33E-01 & \textbf{3.43E-01} \\ \cline{4-11}
      &   & \multirow{3}{*}{$[-10.0, 10.0]$} & \textbf{1.17E+01} & 1.22E+01 & 1.29E+01 & 1.31E+01 & 1.02E+00 & \textbf{6.86E-01} & 9.20E-01 & 8.90E-01 \\
      &   &   & 1.74E+01 & 1.68E+01 & \textbf{1.52E+01} & 1.59E+01 & \textbf{1.56E+00} & 1.94E+00 & 1.57E+00 & 1.78E+00 \\
      &   &   & 1.48E+01 & 1.48E+01 & \textbf{1.37E+01} & 1.45E+01 & 1.31E+00 & 1.39E+00 & \textbf{1.14E+00} & 1.29E+00 \\
    \midrule
    \multirow{12}{*}{Rag} & \multirow{6}{*}{$7 \times 7$} & \multirow{3}{*}{$[-5.0, 5.0]$} & \textbf{6.54E+00} & 6.68E+00 & 7.07E+00 & 7.23E+00 & 1.74E-01 & \textbf{1.59E-01} & 1.70E-01 & 2.02E-01 \\
      &   &   & 1.09E+01 & \textbf{8.72E+00} & 1.04E+01 & 1.08E+01 & 3.16E-01 & \textbf{2.18E-01} & 3.08E-01 & 3.01E-01 \\
      &   &   & 8.38E+00 & \textbf{7.42E+00} & 8.27E+00 & 9.08E+00 & 2.26E-01 & \textbf{1.88E-01} & 2.15E-01 & 2.48E-01 \\ \cline{4-11}
      &   & \multirow{3}{*}{$[-10.0, 10.0]$} & 1.30E+01 & 1.31E+01 & \textbf{1.29E+01} & 1.61E+01 & \textbf{3.62E-01} & 3.68E-01 & 3.84E-01 & 4.70E-01 \\
      &   &   & 1.57E+01 & 1.64E+01 & \textbf{1.53E+01} & 1.85E+01 & 5.19E-01 & 5.22E-01 & \textbf{4.93E-01} & 6.77E-01 \\
      &   &   & \textbf{1.43E+01} & 1.45E+01 & 1.45E+01 & 1.76E+01 & 4.37E-01 & \textbf{4.36E-01} & 4.41E-01 & 5.68E-01 \\ \cline{4-11}
      & \multirow{6}{*}{$11 \times 11$} & \multirow{3}{*}{$[-5.0, 5.0]$} & 1.33E+01 & \textbf{1.15E+01} & 1.29E+01 & 1.37E+01 & 4.61E-01 & \textbf{3.73E-01} & 4.54E-01 & 5.07E-01 \\
      &   &   & 1.74E+01 & \textbf{1.50E+01} & 1.58E+01 & 1.63E+01 & 7.63E-01 & \textbf{5.45E-01} & 5.89E-01 & 6.04E-01 \\
      &   &   & 1.53E+01 & \textbf{1.40E+01} & 1.47E+01 & 1.49E+01 & 5.90E-01 & \textbf{4.87E-01} & 5.27E-01 & 5.43E-01 \\ \cline{4-11}
      &   & \multirow{3}{*}{$[-10.0, 10.0]$} & 1.66E+01 & 1.71E+01 & \textbf{1.55E+01} & 1.89E+01 & 6.91E-01 & 6.42E-01 & \textbf{5.06E-01} & 6.77E-01 \\
      &   &   & 2.66E+01 & 3.00E+01 & 2.86E+01 & \textbf{2.27E+01} & 2.02E+00 & 2.19E+00 & 2.22E+00 & \textbf{9.76E-01} \\
      &   &   & 2.29E+01 & 2.18E+01 & 2.22E+01 & \textbf{2.02E+01} & 1.35E+00 & 1.11E+00 & 1.23E+00 & \textbf{7.90E-01} \\
    \midrule
    \multirow{12}{*}{Alphabet} & \multirow{6}{*}{$7 \times 7$} & \multirow{3}{*}{$[-5.0, 5.0]$} & 5.15E+00 & 5.20E+00 & \textbf{4.78E+00} & 6.62E+00 & 2.03E-01 & 2.13E-01 & \textbf{1.95E-01} & 2.67E-01 \\
      &   &   & 6.23E+00 & \textbf{5.70E+00} & 5.90E+00 & 8.25E+00 & 2.48E-01 & \textbf{2.33E-01} & 2.45E-01 & 3.30E-01 \\
      &   &   & 5.51E+00 & 5.50E+00 & \textbf{5.35E+00} & 7.10E+00 & \textbf{2.21E-01} & 2.21E-01 & 2.24E-01 & 2.88E-01 \\ \cline{4-11}
      &   & \multirow{3}{*}{$[-10.0, 10.0]$} & \textbf{9.87E+00} & 1.09E+01 & 1.06E+01 & 1.25E+01 & \textbf{3.92E-01} & 4.24E-01 & 4.19E-01 & 5.24E-01 \\
      &   &   & \textbf{1.18E+01} & 1.20E+01 & 1.29E+01 & 1.39E+01 & 5.08E-01 & \textbf{4.85E-01} & 5.58E-01 & 6.40E-01 \\
      &   &   & \textbf{1.09E+01} & 1.13E+01 & 1.17E+01 & 1.32E+01 & \textbf{4.65E-01} & 4.67E-01 & 4.96E-01 & 5.72E-01 \\ \cline{4-11}
      & \multirow{6}{*}{$11 \times 11$} & \multirow{3}{*}{$[-5.0, 5.0]$} & 6.93E+00 & 5.76E+00 & \textbf{4.34E+00} & 6.92E+00 & 3.21E-01 & 2.49E-01 & \textbf{1.93E-01} & 3.15E-01 \\
      &   &   & 8.88E+00 & \textbf{6.93E+00} & 1.08E+01 & 8.43E+00 & 3.82E-01 & \textbf{3.18E-01} & 4.62E-01 & 3.96E-01 \\
      &   &   & 7.52E+00 & \textbf{6.59E+00} & 6.82E+00 & 7.65E+00 & 3.49E-01 & \textbf{2.95E-01} & 3.00E-01 & 3.60E-01 \\ \cline{4-11}
      &   & \multirow{3}{*}{$[-10.0, 10.0]$} & 1.84E+01 & 1.19E+01 & 1.08E+01 & \textbf{1.06E+01} & 1.02E+00 & 5.98E-01 & \textbf{5.37E-01} & 5.77E-01 \\
      &   &   & 2.59E+01 & 2.26E+01 & 2.37E+01 & \textbf{1.48E+01} & 1.70E+00 & 1.44E+00 & 1.62E+00 & \textbf{7.67E-01} \\
      &   &   & 2.20E+01 & 1.68E+01 & 1.63E+01 & \textbf{1.26E+01} & 1.29E+00 & 9.69E-01 & 9.61E-01 & \textbf{6.60E-01} \\
    \midrule
    \multirow{12}{*}{Toad} & \multirow{6}{*}{$7 \times 7$} & \multirow{3}{*}{$[-5.0, 5.0]$} & 3.09E+00 & 3.05E+00 & \textbf{2.91E+00} & 3.13E+00 & 2.21E-01 & 2.00E-01 & \textbf{1.82E-01} & 2.29E-01 \\
      &   &   & 3.88E+00 & 3.79E+00 & \textbf{3.42E+00} & 4.33E+00 & 3.52E-01 & 4.26E-01 & \textbf{2.57E-01} & 3.86E-01 \\
      &   &   & 3.62E+00 & 3.40E+00 & \textbf{3.05E+00} & 3.83E+00 & 2.72E-01 & 2.53E-01 & \textbf{2.41E-01} & 3.11E-01 \\ \cline{4-11}
      &   & \multirow{3}{*}{$[-10.0, 10.0]$} & 6.72E+00 & 5.80E+00 & \textbf{4.63E+00} & 7.48E+00 & 5.08E-01 & \textbf{4.70E-01} & 5.02E-01 & 5.93E-01 \\
      &   &   & 7.95E+00 & \textbf{7.14E+00} & 8.59E+00 & 9.89E+00 & 7.05E-01 & 6.39E-01 & \textbf{5.83E-01} & 7.51E-01 \\
      &   &   & 7.48E+00 & 6.48E+00 & \textbf{6.16E+00} & 8.43E+00 & 6.13E-01 & 5.50E-01 & \textbf{5.46E-01} & 6.60E-01 \\ \cline{4-11}
      & \multirow{6}{*}{$11 \times 11$} & \multirow{3}{*}{$[-5.0, 5.0]$} & 3.30E+00 & 2.67E+00 & \textbf{2.51E+00} & 3.36E+00 & 4.48E-01 & \textbf{3.18E-01} & 3.70E-01 & 4.17E-01 \\
      &   &   & 3.91E+00 & \textbf{3.64E+00} & 3.73E+00 & 4.61E+00 & 7.05E-01 & 6.74E-01 & 7.19E-01 & \textbf{5.13E-01} \\
      &   &   & 3.54E+00 & \textbf{3.08E+00} & 3.21E+00 & 3.95E+00 & 6.02E-01 & 5.14E-01 & 4.99E-01 & \textbf{4.61E-01} \\ \cline{4-11}
      &   & \multirow{3}{*}{$[-10.0, 10.0]$} & 7.57E+00 & 6.26E+00 & \textbf{5.95E+00} & 8.72E+00 & 1.35E+00 & 8.29E-01 & 8.65E-01 & \textbf{8.24E-01} \\
      &   &   & 9.98E+00 & 8.55E+00 & \textbf{7.39E+00} & 1.07E+01 & 1.92E+00 & 1.39E+00 & 1.33E+00 & \textbf{1.03E+00} \\
      &   &   & 8.70E+00 & 7.62E+00 & \textbf{6.56E+00} & 9.58E+00 & 1.57E+00 & 1.20E+00 & 1.13E+00 & \textbf{9.22E-01} \\
    \bottomrule
    \end{tabular}
\end{table*}

\subsection{Comparison on Real-World Images}
\label{sec:experiment_real}
\begin{figure*}[!t]
\centering
\subfloat[]{
    \begin{minipage}{0.24\linewidth}
    \centering
        \includegraphics[width=0.5\hsize]{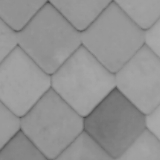}\\
        \includegraphics[width=\hsize]{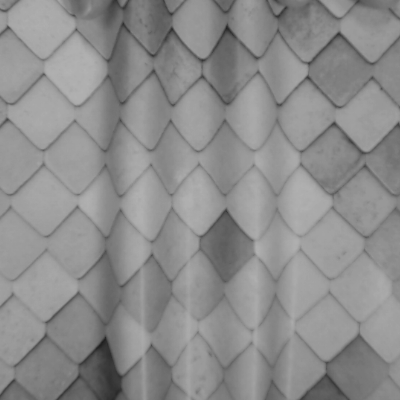}
    \end{minipage}
    \label{fig:real_image_texture}
}\quad
\subfloat[]{
    \begin{minipage}{0.24\linewidth}
    \centering
        \includegraphics[width=0.5\hsize]{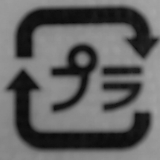}\\
        \includegraphics[width=\hsize]{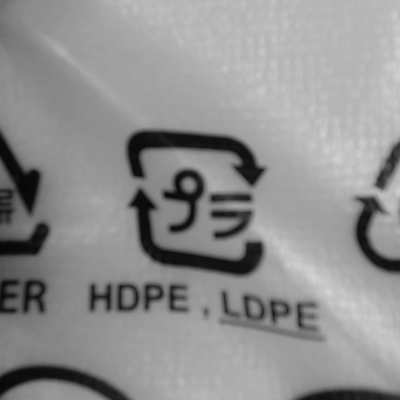}
    \end{minipage}
    \label{fig:real_image_sign}
}\quad
\subfloat[]{
    \begin{minipage}{0.24\linewidth}
    \centering
        \includegraphics[width=0.5\hsize]{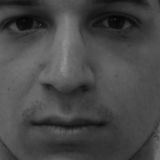}\\
        \includegraphics[width=\hsize]{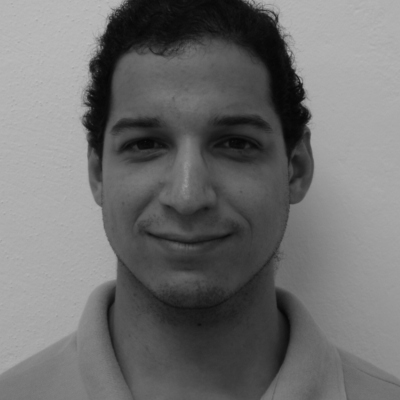}
    \end{minipage}
    \label{fig:real_image_face}
}
\caption{Real-world template images (top row) and target images (bottom row). (a) texture, (b) sign, and (c) face.}
\label{fig:real_image}
\end{figure*}

The usefulness of the proposed method under real-world scenarios is verified in this section. We use three different pairs of template and target images as shown in Fig. \ref{fig:real_image}.
\begin{itemize}
    \item Texture (Fig. \ref{fig:real_image_texture}): the images capture a part of the undeformed/deformed texture printed on a piece of paper. The target image has two vertical bumps. We set the lattice as $7 \times 7$  and the decision variable range as $[-20.0, 20.0]$.
    \item Sign (Fig. \ref{fig:real_image_sign}): the images capture a sign with undeformed/deformed text printed on a piece of wrapping paper. The lattice and decision variable range are set to $7 \times 7$ lattice and $[-10.0, 10.0]$, respectively.
    \item Face (Fig. \ref{fig:real_image_face}): the template image and the target image show a frontal face with serious expression and smile, respectively. The goal is to obtain deformation parameters that express smiling. We use the $7 \times 7$ lattice and the $[-20.0, 20.0]$ decision variable range.
\end{itemize}
The sizes of the template images and target images are set the same as Sec. \ref{sec:experiment_synthetic}. Fig. \ref{fig:real_image_texture} and Fig. \ref{fig:real_image_sign} are captured by a web camera and Fig. \ref{fig:real_image_face} is extracted from the FEI face database\footnote{\url{https://fei.edu.br/~cet/facedatabase.html}} \cite{thomaz2010new} and cropped. Because the ground truth of the real deformations is unknown, we evaluate results only using RMSE.

\begin{table*}[!t]
\centering
\caption{The RMSE results of best solutions and post-processed solutions with respect to the \textit{real-world images}. Each cell consists of three values which represent the minimum, maximum, and average values based on five random trials from top to bottom. The values of post-processed solutions which are smaller than the values of the corresponding best solutions are shown in Italics. The best results in terms of RMSE with respect to each row are emphasized in bold.}
\label{tab:real_image}
\scriptsize
    \begin{tabular}{cccrrrrrrr}
    \toprule
      &   &   & \multicolumn{4}{c}{Best solution} & \multicolumn{3}{c}{Post-processed solution} \\
    \multirow{3}{*}{Image} & \multirow{3}{*}{\begin{tabular}{c} Lattice \\ size \end{tabular}} & Decision & \multicolumn{1}{c}{\multirow{3}{*}{GA}} & \multicolumn{1}{c}{Two} & \multicolumn{1}{c}{Two} & \multicolumn{1}{c}{Four} & \multicolumn{1}{c}{Two} & \multicolumn{1}{c}{Two} & \multicolumn{1}{c}{Four} \\
      &   & variable &   & \multicolumn{1}{c}{objective} & \multicolumn{1}{c}{objective} & \multicolumn{1}{c}{objective} & \multicolumn{1}{c}{objective} & \multicolumn{1}{c}{objective} & \multicolumn{1}{c}{objective} \\
      &   & range &   & \multicolumn{1}{c}{NSGA-II} & \multicolumn{1}{c}{NSGA-III} & \multicolumn{1}{c}{NSGA-III} & \multicolumn{1}{c}{NSGA-II} & \multicolumn{1}{c}{NSGA-III} & \multicolumn{1}{c}{NSGA-III} \\
    \midrule
    \midrule
    \multirow{3}{*}{Texture} & \multirow{3}{*}{$7 \times 7$} & \multirow{3}{*}{$[-20.0, 20.0]$} & 1.23E+01 & 1.21E+01 & 1.24E+01 & \textbf{1.18E+01} & 1.23E+01 & \textit{1.23E+01} & 1.19E+01 \\
      &   &   & 1.29E+01 & \textbf{1.25E+01} & 1.28E+01 & 1.25E+01 & 1.25E+01 & 1.28E+01 & \textit{1.25E+01} \\
      &   &   & 1.25E+01 & 1.23E+01 & 1.25E+01 & \textbf{1.21E+01} & 1.24E+01 & 1.26E+01 & 1.22E+01 \\
    \midrule
    \multirow{3}{*}{Sign} & \multirow{3}{*}{$7 \times 7$} & \multirow{3}{*}{$[-10.0, 10.0]$} & 2.04E+01 & 1.95E+01 & 1.95E+01 & 2.04E+01 & 1.98E+01 & \textit{\textbf{1.95E+01}} & \textit{2.02E+01} \\
      &   &   & 2.13E+01 & \textbf{2.03E+01} & 2.06E+01 & 2.07E+01 & 2.04E+01 & 2.10E+01 & \textit{2.06E+01} \\
      &   &   & 2.09E+01 & 2.00E+01 & \textbf{2.00E+01} & 2.05E+01 & 2.01E+01 & 2.00E+01 & \textit{2.05E+01} \\
    \midrule
    \multirow{3}{*}{Face} & \multirow{3}{*}{$7 \times 7$} & \multirow{3}{*}{$[-20.0, 20.0]$} & 7.55E+00 & 7.42E+00 & \textbf{7.37E+00} & 7.50E+00 & 7.47E+00 & 7.42E+00 & 7.50E+00 \\
      &   &   & 8.01E+00 & 7.71E+00 & \textbf{7.61E+00} & 7.87E+00 & 7.88E+00 & 7.82E+00 & 8.08E+00 \\
      &   &   & 7.72E+00 & 7.58E+00 & \textbf{7.48E+00} & 7.69E+00 & 7.65E+00 & 7.56E+00 & 7.77E+00 \\
    \bottomrule
    \end{tabular}
\end{table*}

\begin{figure*}[!t]
\centering
\subfloat[texture]{
    \begin{minipage}{\linewidth}
    \centering
        \includegraphics[width=0.24\hsize]{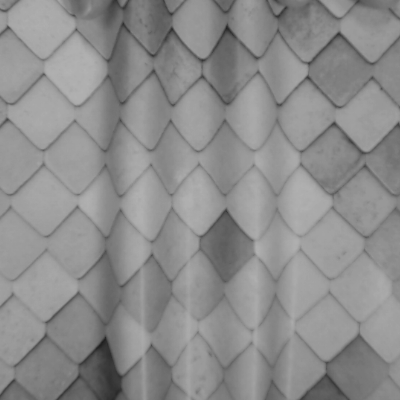}
        \includegraphics[width=0.24\hsize]{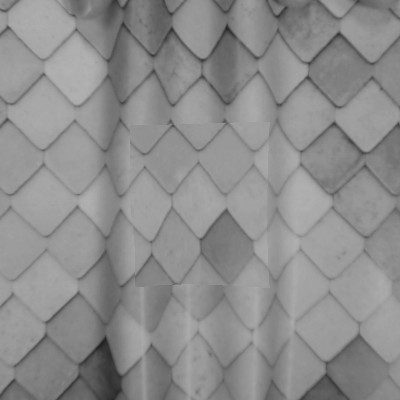}
        \includegraphics[width=0.24\hsize]{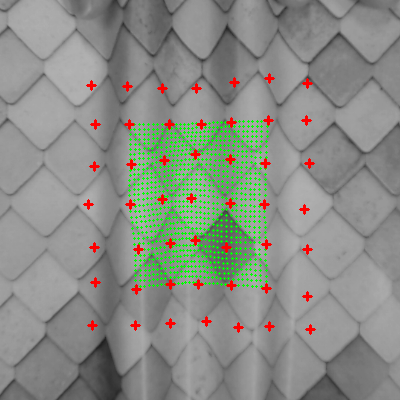}
        \includegraphics[width=0.24\hsize]{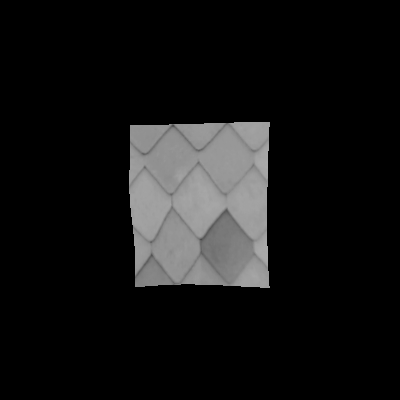}
    \end{minipage}
    \label{fig:real_success_texture}
}\\
\subfloat[sign]{
    \begin{minipage}{\linewidth}
    \centering
        \includegraphics[width=0.24\hsize]{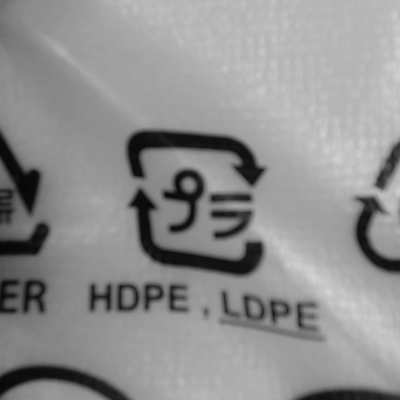}
        \includegraphics[width=0.24\hsize]{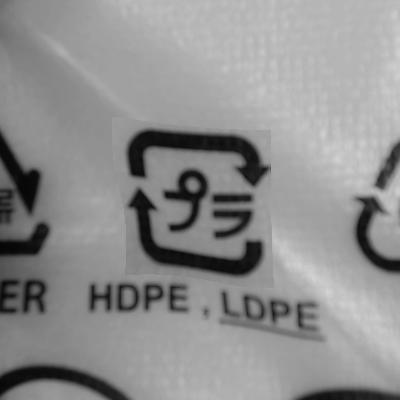}
        \includegraphics[width=0.24\hsize]{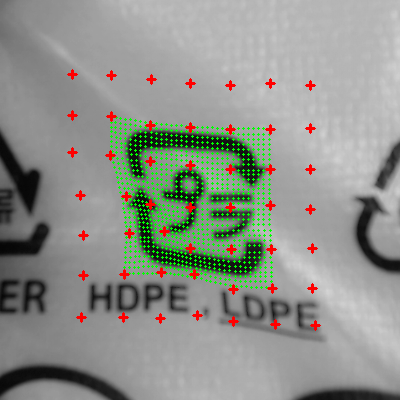}
        \includegraphics[width=0.24\hsize]{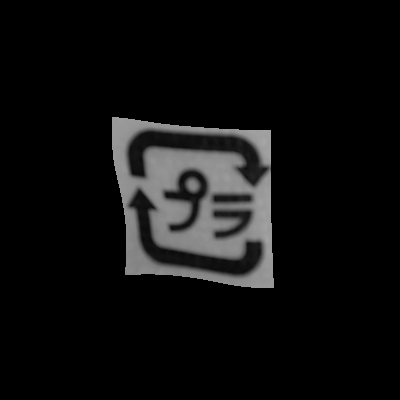}
    \end{minipage}
    \label{fig:real_success_sign}
}\\
\subfloat[face]{
    \begin{minipage}{\linewidth}
    \centering
        \includegraphics[width=0.24\hsize]{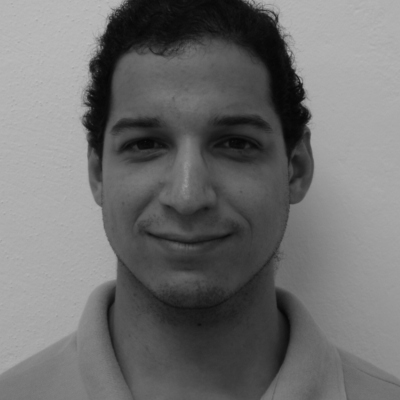}
        \includegraphics[width=0.24\hsize]{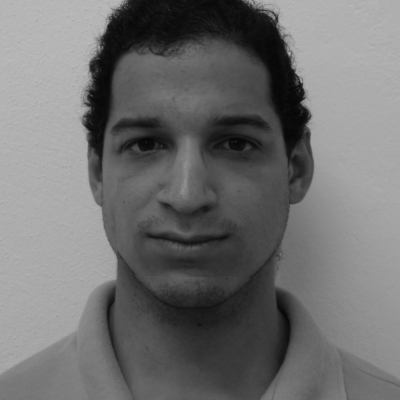}
        \includegraphics[width=0.24\hsize]{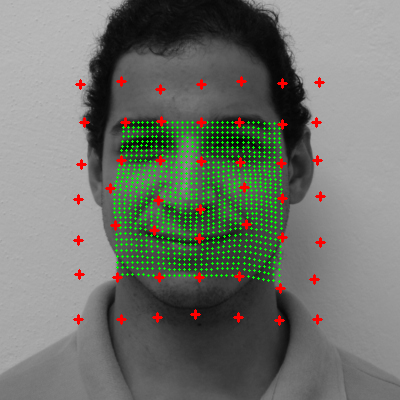}
        \includegraphics[width=0.24\hsize]{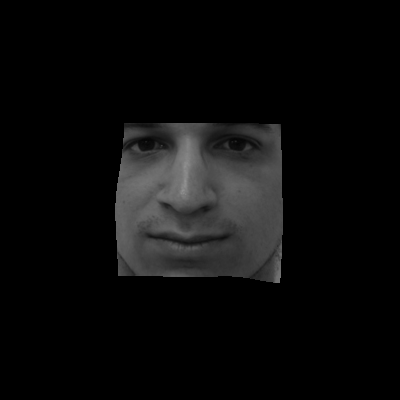}
    \end{minipage}
    \label{fig:real_success_face}
}
\caption{Target images (1st column) and estimation results (2nd$\sim$4th columns). The 4th column shows the template deformed with estimated deformation. Composite images in the 2nd column are created from the images in the 1st column and 4th column.}
\label{fig:real_success}
\end{figure*}

The results of best solutions and post-processed solutions are shown in Table \ref{tab:real_image}. These results show that multi-objective approaches can outperform the single-objective approach in real-world situations. However,  post-processing fails to improve estimation accuracy in a number of cases (e.g., the face image). The estimation results for each image are shown in Fig. \ref{fig:real_success}. It can be observed that the deformed texture image contains some highlight areas and the eye area of the face image is also shadowed. Hence, the deformation results for these corresponding areas show worse accuracy than the other areas. The groups including these areas cannot contribute to the post-processing. As a limitation, the proposed method suffers from illumination changes due to the characteristics of the similarity measure.

\section{Conclusion}
\label{sec:conclusion}
In this paper, we proposed a novel deformation estimation method using MOEAs to tackle the conflicts based on the fact that each control point of the deformation model affects a local region rather than a single pixel. Our method casts deformation estimation as a MOP by dividing a template image into several groups consisting of patches with group-wise similarity defined as group-wise objective functions, which can be solved by off-the-shelf MOEAs. To handle large deformations, optimization is run hierarchically following a coarse-to-fine strategy powered by image pyramid and control point mesh subdivision. Besides, a post-processing procedure is proposed to integrate Pareto optimal solutions into a single output, which can improve the estimation accuracy. The observations from experimental results can be summarized threefold. First, our partitioning approach with two-objective algorithms can obtain deformation parameters more accurately than GA with a single objective. Second, although the four-objective algorithm performs not as well as expected due to a large number of objectives, it shows to be effective in dealing with complex and subtle deformations. Third, the post-processing procedure can improve estimation accuracy in many cases. We can observe the usefulness of the proposed method with real-world images.

The main limitation of our method is that high computational resources are required. As future work, we would like to address this issue by further tuning the hyper-parameters that can reduce the computational cost without degrading the performance. We are also interested in the referenced point distribution of the NSGA-III. A user-supplied setting may be able to focus solutions on regions that are desirable for the post-processing procedure.

\bibliographystyle{IEEEtran}
\bibliography{references}

\end{document}